%% file: neurips_2024.tex
\documentclass{article}

\usepackage[preprint]{neurips_2024}

\usepackage[utf8]{inputenc} 
\usepackage[T1]{fontenc}    
\usepackage{hyperref}       
\usepackage{url}            
\usepackage{booktabs}       
\usepackage{amsfonts}       
\usepackage{nicefrac}       
\usepackage{microtype}      
\usepackage{xcolor}         

\usepackage{graphicx}
\usepackage{pifont}
\usepackage{subcaption}
\usepackage{placeins}
\usepackage{multirow}
\usepackage{multicol}
\usepackage{amsmath} 
\usepackage{cleveref}
\usepackage{algorithm}
\usepackage{algorithmic}
\usepackage{xcolor}
\usepackage{makecell}
\usepackage{comment}
\usepackage{lineno}
\usepackage{listings}

\crefname{table}{Figure}{Figures}
\crefname{figure}{Figure}{Figures}
\crefname{appendix}{Appendix}{}
\crefname{section}{Section}{}

\makeatletter
\let\c@table\c@figure     
\makeatother

\newcommand{\cmark}{\ding{51}}  
\newcommand{\xmark}{\ding{55}}  

\title{Scaling Up Forest Vision with Synthetic Data}

\author{%
  Yihang She\\
  University of Cambridge\\
  \texttt{ys611@cam.ac.uk} \\
  \And
  Andrew Blake \\
  University of Cambridge \\
  \texttt{ab@ablake.ai} \\
  \And
  David Coomes \\
  University of Cambridge \\
  \texttt{dac18@cam.ac.uk} \\
  \And
  Srinivasan Keshav \\
  University of Cambridge \\
  \texttt{sk818@cam.ac.uk} \\
}

\begin{document}

\maketitle

\begin{abstract}
\input{0_abstract}

\end{abstract}

\input{1_introduction}

\input{2_related_work}

\input{3_methods}
\input{4_results}

\input{5_conclusion}

\clearpage
\bibliographystyle{unsrtnat}
\bibliography{neurips_2024}

\newpage
\appendix
\input{6_appendix}


\end{document}

%% file: 0_abstract.tex
Accurate tree segmentation is a key step in extracting individual tree metrics from forest laser scans, and is essential to understanding ecosystem functions in carbon cycling and beyond.
Over the past decade, tree segmentation algorithms have advanced rapidly due to developments in AI. However existing, public, 3D forest datasets are not large enough to build robust tree segmentation systems. 
Motivated by the success of synthetic data in other domains such as self-driving, we investigate whether similar approaches can help with tree segmentation.
In place of expensive field data collection and annotation, we use synthetic data during pretraining, and then require only minimal, real forest plot annotation for fine-tuning.

We have developed a new synthetic data generation pipeline to do this for forest vision tasks, integrating advances in game-engines with physics-based LiDAR simulation. 
As a result, we have produced a comprehensive, diverse, annotated 3D forest dataset on an unprecedented scale.
Extensive experiments with a state-of-the-art tree segmentation algorithm and a popular real dataset show that our synthetic data can substantially reduce the need for labelled real data.
After fine-tuning on just a single, real, forest plot of less than 0.1 hectare, the pretrained model achieves segmentations that are competitive with a model trained on the full scale real data. 
We have also identified critical factors for successful use of synthetic data: physics, diversity, and scale, paving the way for more robust 3D forest vision systems in the future.
Our data generation pipeline and the resulting dataset are available at \url{https://github.com/yihshe/CAMP3D.git}.

%% file: 1_introduction.tex
\section{Introduction}\label{sec:intro}
Understanding forest dynamics and monitoring their status have long been a priority in the Sustainable Development Goals of the United Nations \citep{swamy2018future}. Forests can be monitored at different spatial scales from ground surveys to space missions, with varying trade-offs between accuracy, cost, and spatial coverage \citep{lines2022shape,liang2022close,dubayah2020global}. Forest monitoring at a close range \citep{liang2022close} has become increasingly important for precise forest management. It also offers ground reference data to calibrate satellite remote sensing \citep{lang2022global}. 

\begin{figure}[htbp]
    \centering
    \includegraphics[width=\textwidth]{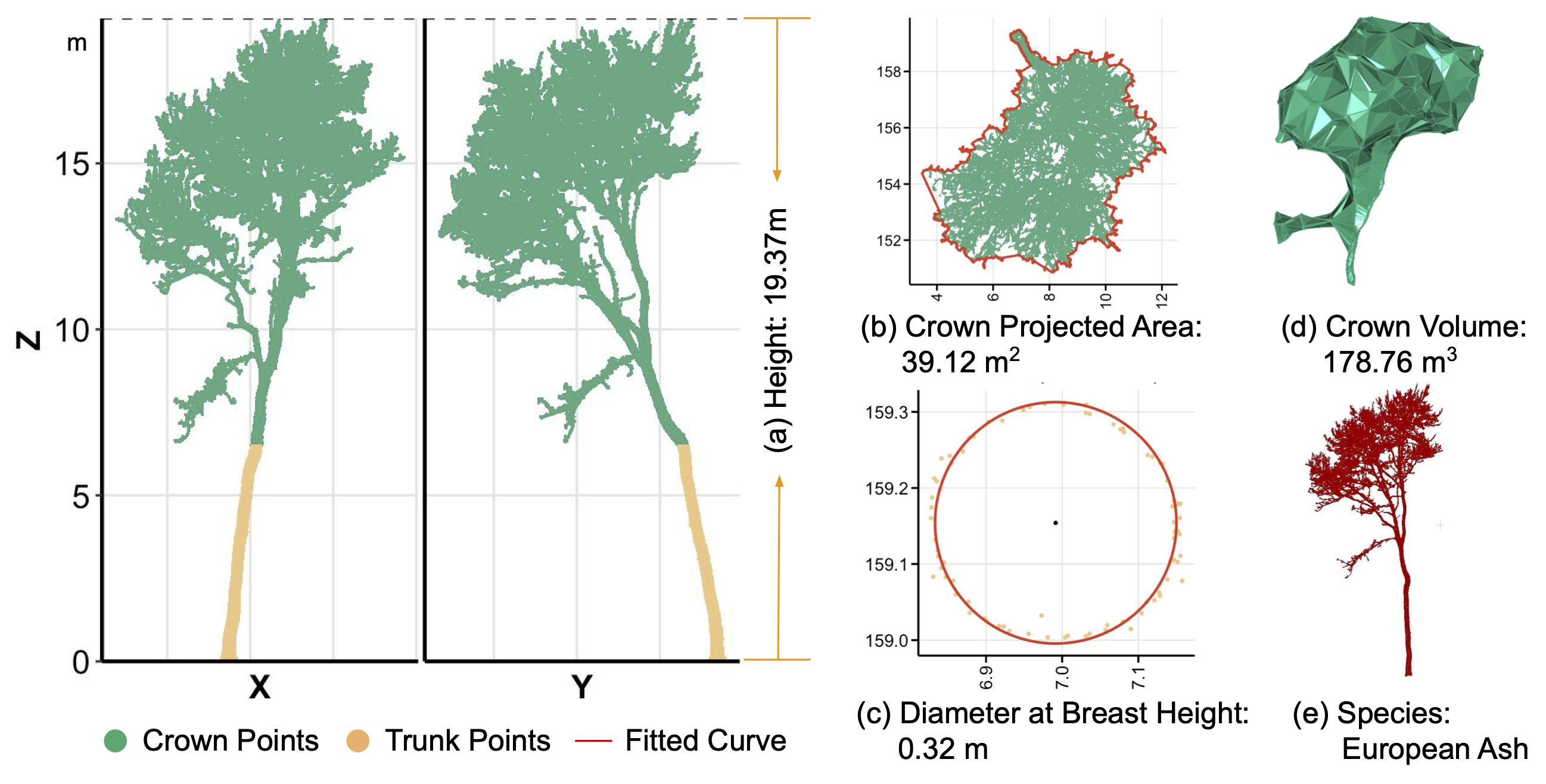}
    \caption{\textbf{Individual tree segmentation is essential for deriving key forest properties in ecology.} \textit{This figure shows examples of metrics derived from individual tree segmentation: canopy height and diameter at breast height for biomass and carbon estimation, and crown area, crown volume, and species composition for understanding forest structure, biodiversity, and ecosystem functions.}}
    \label{fig:methods_tree_properties}
\end{figure}

For close-range forest sensing to obtain metrics for individual trees, the typical workflow involves obtaining point clouds of forests through laser scanning and segmenting individual trees. 
The next step is to extract the properties of interest, such as above-ground biomass \citep{coomes2017area}, species \citep{puliti2024benchmarking}, tree health \citep{chan2021monitoring}, crown volume, canopy height and diameter at breast height (DBH) \citep{liang2019forest}, among others. 
Accurate measurement of these properties is essential for understanding key ecosystem functions related to carbon cycling and beyond (\cref{fig:methods_tree_properties}). For example, canopy height and DBH are fundamental for quantifying forest metabolic scaling and allometry, which underpin the estimation of above-ground biomass \citep{chave2009towards}. The shape of the canopy surface and the volume of the crown reveal competitive light dynamics, providing indicators of tree health and reflecting the role of the forest in the energy cascade of ecosystems \citep{purves2007crown}. Metrics derived from canopy surfaces are also critical for studying canopy–atmosphere gas exchange, including the uptake and release of greenhouse gases such as methane and carbon dioxide \citep{gauci2024global}.
The accurate segmentation of individual trees is a key bottleneck in the entire pipeline.

The choice of algorithm plays a crucial role in building a robust tree segmentation system. Over the past decade, tree segmentation has changed from the use of the classical region-growing algorithm \citep{coomes2017area} to increasingly complicated neural network architectures \citep{straker2023instance,ball2022accurate,wielgosz2023point2tree,xiang2024automated,wielgosz2024segmentanytree}. It also expanded from 2D image operations \citep{coomes2017area,ball2022accurate,straker2023instance} to the processing of 3D point clouds \citep{wielgosz2023point2tree,xiang2024automated,wielgosz2024segmentanytree}. All of this progress has refined the segmentation of trees for a more accurate retrieval of forest properties as discussed above. 

The increasing complexity of tree segmentation algorithms also requires more data and computation for a stronger and more generalisable model performance \citep{sutton2019bitter}. Despite extensive data collection efforts \citep{weiser2022individual,calders2022laser,puliti2023instance}, the available 3D forest dataset for the development of tree segmentation algorithms still falls short of the scale typically required for robust machine learning \citep{uddin2024dataset}. This is not surprising given that forest scanning requires field work that is limited by multiple factors, such as site accessibility, weather conditions, and seasons. In addition, the intricate canopy structures of forests make it challenging and time consuming to delineate the semantic labels (classifying each point by category, e.g., stem, branch, ground) and instance labels (assigning unique identifiers to individual trees). For example, \citet{puliti2023instance} report that just annotating 2.79 hectares of forest plots takes two skilled researchers six months. \citep{wielgosz2024segmentanytree} suggest that the scarcity of data has become a bottleneck for further improvement of algorithm performance.

Recent advances in graphics technology enable synthetic data generation to tackle data scarcity in other vision domains \citep{shotton2011real,sklyarova2023haar,gaidon2016virtual,roberts2021hypersim}. In self-driving vehicles, the use of synthetic data has become a key factor for the rapid development of this field \citep{song2023synthetic}. In the context of forest monitoring, procedural foliage generation in graphics engines such as Unreal Engine (UE) has allowed the creation of forest scenes on a large scale \citep{holmberg2025unreal}. Researchers have repurposed  game scenes to render synthetic 2D images for the development of tree segmentation algorithms \citep{Grondin_2022,lu2024m2fnet,feng2025spread}. Open source graphics engines such as Blender enable the creation of simulation packages tailored to natural scenes \citep{liu2021infinite,winiwarter2022virtual}. Simulators such as the Heidelberg LiDAR Operations Simulator (HELIOS) \citep{winiwarter2022virtual} resemble the real-world laser scan process during a forest survey. 

Synthetic data can significantly reduce the effort to collect data in the field and then to annotate the data, as many existing synthetic forest datasets have demonstrated \citep{Grondin_2022,lu2024m2fnet,wang2020unsupervised,bryson2024domain}. Nevertheless, its effectiveness depends on whether an algorithm trained on the synthetic data can generalise well to the real world. A key factor is the diversity of synthetic data on large scales: with enough diversity, synthetic data could allow the algorithm trained on it to generalize to real-world forests \citep{tobin2017domain,prakash2019structured}, and with large-scale data, neural networks can benefit from adequate computations for effective weight and bias learning \citep{sutton2019bitter}. Under this assumption, the learning paradigm for a forest vision algorithm would involve pretraining on synthetic data, followed by fine-tuning on real data from the region of interest. This approach, similar to meta-learning \citep{finn2017model}, is expected to enable the pretrained algorithm to generalize effectively to real-world scenarios, even when only a limited number of labelled samples are available.

Although several synthetic forest datasets have been developed using advanced simulation tools such as HELIOS \citep{winiwarter2022virtual}, most are of relatively small scale and limited in structural diversity. For example, \citet{wang2020unsupervised} manually created a 0.16 ha plot (40 m $\times$ 40 m) containing ~30 trees of a single species, while \citet{bryson2024domain} generated a plot of ~400 trees in a regularly spaced, low-variability layout. The recently introduced Boreal3D dataset \citep{liu2025advancing} is the largest to date, with 1,000 plots (each 20 m × 20 m) derived from real-world TLS scans of six coniferous plots in Finland--- totalling 40 ha but covering only three coniferous species.

In contrast, our approach sources forest scenes from the gaming industry and regenerates them using procedural foliage generation, originally developed in Unreal Engine but, as we find, grounded in ecological forest-growth principles. This enables the creation of structurally diverse forest scenes by leveraging prebuilt forest scenes and tree assets, repurposed to simulate point clouds for forest vision tasks. We then integrate these scenes with the HELIOS simulator to produce physically-based point clouds with accurate semantic and instance labels. The resulting data set spans 12 forest scenes over 75 ha, covering both coniferous and deciduous types, and exceeds the scale and diversity of real dataset such as Wytham Woods (1.4 ha) \citep{calders2022laser} and FOR-Instance (2.79 ha) \citep{puliti2023instance} and synthetic dataset such as Boreal3D (40 ha, limited diversity) \citep{liu2025advancing}.

Using our synthetic data, we experiment with the learning paradigm of pretraining followed by fine-tuning. We pre-train a state-of-the-art 3D tree segmentation algorithm \citep{xiang2024automated} on our synthetic dataset and validate it on a widely used real-world data set for tree segmentation \citep{puliti2023instance}. Experiments show that the algorithm pretrained with our synthetic data gets significant improvement than training from scratch on a few real plots. Fine-tuning the pretrained algorithm on just one forest plot achieves segmentation accuracy comparable to full-scale real data, although this plot can represent as little as 2.2\% of the labelled dataset.
In summary, our contributions include: 
\begin{itemize}
    \item a synthetic data generation pipeline for forest vision tasks that combine  advances in game engines and physics-based simulation; 
    \item a comprehensive, diverse, and large-scale synthetic dataset of 3D point-clouds  for tree segmentation tasks;
    \item experiments showing the effectiveness of the synthetic dataset, and revealing that physics-based simulation, scene diversity, and large dataset scale are all crucial for synthetic data to be effective for training.
\end{itemize}

%% file: 2_related_work.tex
\section{Related Work}
\subsection{Close-range sensing of forest}
Close-range sensing in forest monitoring involves using sensors from a few to several hundred meters away to gather detailed, contact-free data on trees and vegetation \citep{liang2022close}.
Compared to remote sensing, close-range sensing of forest offers more fine-grained monitoring of forest status \citep{liang2022close,lang2023high}. It covers a wide range of platforms and sensors from laser scanners to RGB cameras, resulting in different data representations \citep{okura20223d}. In terms of sensor, laser scans have become a dominating sensing method in this field, and point clouds are the most common data representations. These sensing methods can be attributed to different platforms, such as terrestrial laser scanning (TLS) \citep{lines2022shape}, airborne laser scanning (ALS) \citep{coomes2017area}, and unmanned aerial vehicle laser scanning (UAV-LS) \citep{nesbit2019enhancing}. Each platform presents varying levels of trade-offs between accuracy and cost. TLS offers fine-grained information about tree structures; however, it has low spatial coverage and is expensive to operate \citep{lines2022shape}. ALS can cover relatively large areas, but has low resolution and accuracy \citep{coomes2017area}. Recently, UAV advancements have provided a low-cost, high-resolution alternative to terrestrial and airborne platforms, gaining popularity in forest monitoring \citep{nesbit2019enhancing, ball2022accurate, flynn2024technological}.
Accurate segmentation of individual trees is the key step to obtaining the properties of interest at an individual tree level \citep{coomes2017area,liang2019forest,chan2021monitoring,puliti2024benchmarking}. 

\subsection{Tree segmentation algorithms and datasets}
Tree segmentation algorithms have made outstanding progress in recent years due to the fast development of AI \citep{straker2023instance,wielgosz2023point2tree,xiang2023towards,xiang2024automated,wielgosz2024segmentanytree}. Compared to segmenting trees from 2D images \citep{ball2022accurate,straker2023instance}, segmenting trees from 3D point clouds can offer more accurate results, especially in dense canopies \citep{xiang2024automated}. Moreover, 3D segmentation enables a wider range of tasks such as estimating crown volume using point clouds of individual trees instead of just their detected tree crowns \citep{liang2018international,liang2019forest}. In terms of 3D tree segmentation, the state-of-the-art is ForAINet, developed by \citet{xiang2024automated}. It performs sparse 3D convolution on point clouds and has multiple task heads designed for both instance and semantic segmentation. SegAnyTree \citep{wielgosz2023point2tree} employs the same architecture, but generalises the algorithm to different sensors by incorporating UAV and mobile laser scans, and downsampling dense point clouds to simulate airborne laser scans.
The development and evaluation of algorithms are contingent on 3D forest point clouds with quality annotations. Only a few of such datasets are publicly available \citep{calders2022laser,puliti2023instance,weiser2022individual}. For example, the Wytham Woods dataset \citep{calders2016large} covers approximately 1 hectare of TLS scans of deciduous forests in southern England with instance segmentation labels. The FOR instance data set \citep{puliti2023instance} is a UAV-LS data set that covers five regions with instance and semantic segmentation labels, totalling 2.79 ha. In fact, both ForAINet \citep{xiang2024automated} and SegAnyTree \citep{wielgosz2024segmentanytree} are developed using the FOR-instance dataset \citep{puliti2023instance}.

\subsection{Synthetic data and its applications}
Synthetic data is extensively used in fields like human pose estimation \citep{shotton2011real,sklyarova2023haar}, self-driving vehicles \citep{gaidon2016virtual}, and indoor scene understanding \citep{roberts2021hypersim} to combat real-world data scarcity.
Its success is benefited from recent advances in graphics engines and simulators built on top of them \citep{dosovitskiy2017carla,bondi2018airsim,liu2021infinite,winiwarter2022virtual}. In particular, Unreal Engine (UE) offers advanced rendering and procedural foliage generation for easy creation of large-scale forest scenes \citep{holmberg2025unreal}. Synthetic data sets have been created for 2D tree segmentation based on this feature \citep{Grondin_2022, lu2024m2fnet, feng2025spread}. Using UE and AirSim simulators \citep{bondi2018airsim}, \citet{feng2025spread} created the Synthetic Photorealistic Arboreal Dataset (SPREAD) with various forest scenes for training the 2D trunk segmentation algorithm.  
In parallel to UE-based simulation, some simulators have been built on Blender, which is open source and more customisable. Infinigen \citep{liu2021infinite} focuses on rendering synthetic data for natural scenes in Blender, mainly for 2D scenarios. HELIOS \citep{winiwarter2022virtual} is a simulator specifically designed for cross-platform forest laser scanning, meeting our requirements. 
This work introduces a novel data pipeline bridging UE's advanced scene generation and HELIOS's 3D LiDAR forest survey simulation.
While we focus on generating synthetic data to improve machine learning algorithms, existing work has also used synthetic data for other studies of forests. These include comparing simulations of different sensor types with real-world laser scans \citep{esmoris2024deep, schafer2023generating} and simulating the forest radiative transfer process \citep{liu2022implications}. In some of the studies available that generate synthetic laser scans for training tree segmentation algorithms \citep{wang2020unsupervised, bryson2024domain}, forest stands are typically built manually. These are available on a small scale, such as a virtual 40 m * 40 m plot with 30 trees \citep{wang2020unsupervised}.

%% file: 3_methods.tex
\section{Methods}
\subsection{Synthetic data generation pipeline}
\begin{figure}[htbp]
    \centering
    \includegraphics[width=\textwidth]{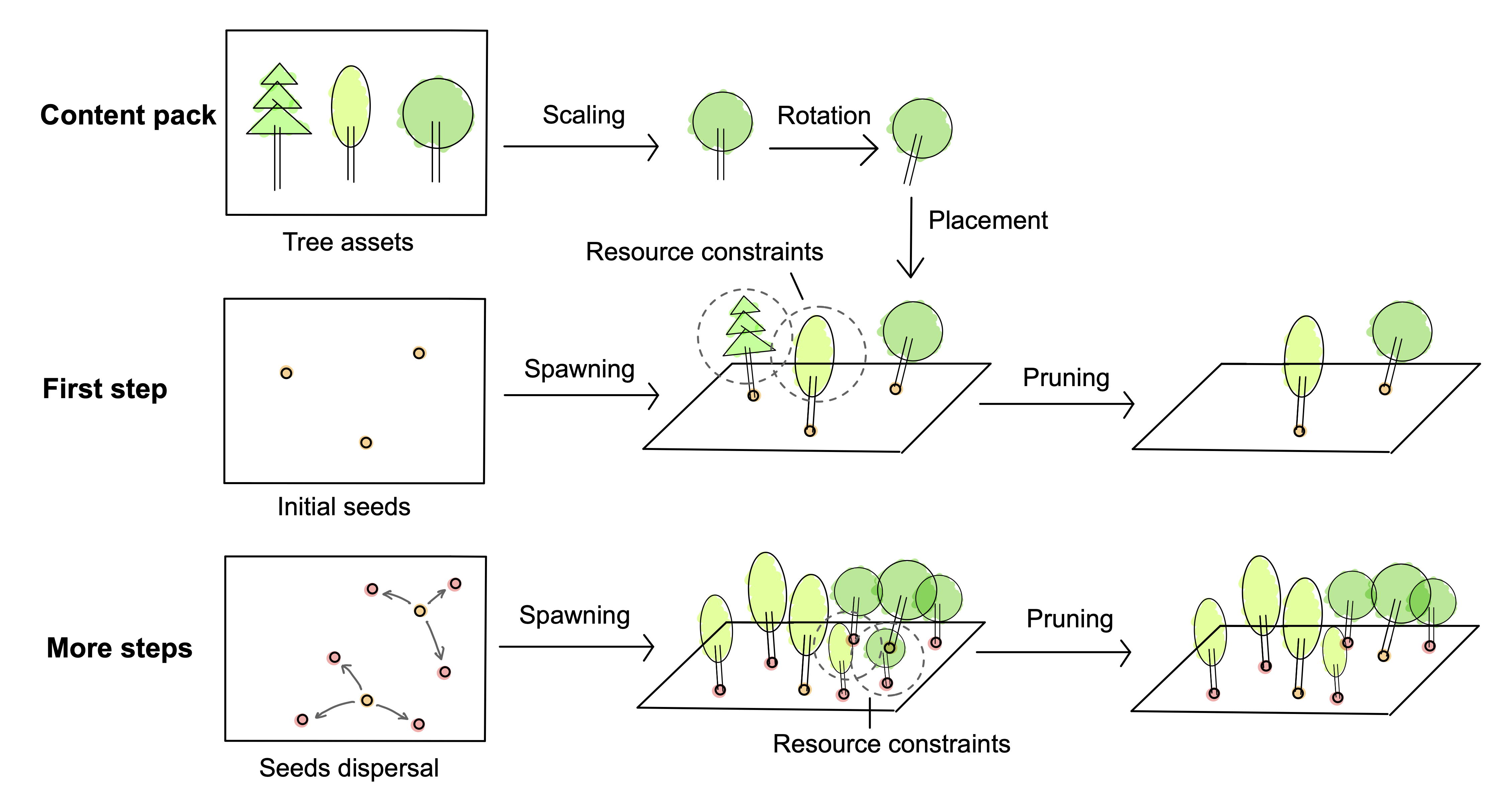}
    \caption{\textbf{Photorealistic forest scenes can be generated, at large scale, with a limited set of tree models, using procedural foliage generation.} \textit{All forest scenes we have collected from Unreal Engine were generated using this algorithm. Although developed and used mainly in the game industry, this algorithm has  roots in ecological theory on forest growth \citep{pacala1996forest,purves2007crown}.}}
    \label{fig:methods_procedural_foliage_generation}
\end{figure}

\textbf{Procedural foliage generation.} We generate forest scenes for simulation using Unreal Engine's procedural foliage generation algorithm ($\mathcal{A}_{\text{PFG}}$; see \cref{fig:methods_procedural_foliage_generation}), which enables large-scale 3D forest generation from a limited set of foliage models. 
While the Unreal Engine community provides many pre-built scenes and asset packs, these are designed mainly for gaming applications and often contain non-forest elements or unsuitable layouts. We therefore regenerate the forest component within a predefined 250,m $\times$ 250,m plot, applying $\mathcal{A}_{\text{PFG}}$ with only tree foliage generators; if only tree assets are provided, we first define the foliage generators from these assets in order to run $\mathcal{A}_{\text{PFG}}$.

This procedural foliage generation algorithm ($\mathcal{A}_{\text{PFG}}$), despite having originated from the game industry, is broadly consistent with ecological theory of forests originating and forming canopy structure through light and resource competition \citep{halle1970essay,pacala1996forest,purves2007crown}: given a bare plot of land, starting from seeds, trees grow and expand to fill all available space, but subject to constraints on light, represented by collision and shade radius in this algorithm. After enough years, or simulation steps, the scene reaches a state of equilibrium --- a closed canopy --- while still respecting individual space requirements. 
At a high level, the algorithm iteratively spawns and prunes foliage in simulation steps. \cref{tab:procedural_paras_coniferous} summarises some key parameters for procedural generation and examples of parameter ranges that can be used for coniferous forests. More details regarding these parameters can be found in \cref{sec:procedural_parameters_appx}.

\begin{table}[htbp]
\centering
\caption{\textbf{Procedural Foliage Parameters} for Coniferous Forests. \textit{Most of these parameters have physical meanings regarding how seeds of a tree can disperse and how the corresponding trees can be spawned.}}
\resizebox{\textwidth}{!}{
\begin{tabular}{lcc}
\toprule
\textbf{Parameter} & \textbf{Overstory (Canopy Trees)} & \textbf{Understory (Saplings/Shrubs)} \\
\midrule
Initial Seed Density        & 3.0         & 2.0         \\
Collision Radius            & 250--300 cm & 50--100 cm  \\
Shade Radius                & 400--500 cm & 50 cm       \\
Procedural Scale            & 0.8 -- 1.2  & 0.5 -- 1.0  \\
Average Spread Distance     & 1500 cm     & 800 cm      \\
Spread Variance             & 500 cm      & 300 cm      \\
Num Steps                   & 2           & 2           \\
Max Age                     & 3           & 2           \\
\bottomrule
\end{tabular}
}
\label{tab:procedural_paras_coniferous}
\end{table}

With our generation method established, we then select and adapt forest scenes from available datasets and asset packs to build a diverse, simulation-ready collection.
We first adapt five natural forest scenes from the SPREAD dataset \citep{feng2025spread}, originally developed for generating 2D images for trunk segmentation. The dataset contains six natural forests, four urban tree scenes, and one plantation; one natural forest scene (``Burned Wood'') is excluded, leaving three deciduous forests, one redwood forest, and one rainforest. In its original form, each SPREAD forest scene covers a 1000 m $\times$ 1000 m area constructed from only ten unique 100 m $\times$ 100 m tiles---a common default setting in game scene generation. While adequate for gaming applications, this design produces repeated spatial patterns, leading to redundancy in simulated data. To avoid this limitation, we regenerate each selected scene in a 250,m $\times$ 250,m (6.25 ha) plot with unique layouts containing only tree vegetation.

Because the SPREAD collection lacks European coniferous forests and has limited European deciduous coverage, we supplement it with seven additional forest scenes originally from the Unreal Engine Marketplace. These scenes are selected for their ecological relevance, specifically European coniferous and deciduous forests, and for the availability of high-quality tree assets at the time of collection. No methodological limit is imposed, so more scenes can be incorporated as suitable assets become available. In total, the dataset comprises twelve adapted forest scenes, covering 75 ha.

\begin{figure}[htbp]
    \centering
    \includegraphics[width=\textwidth]{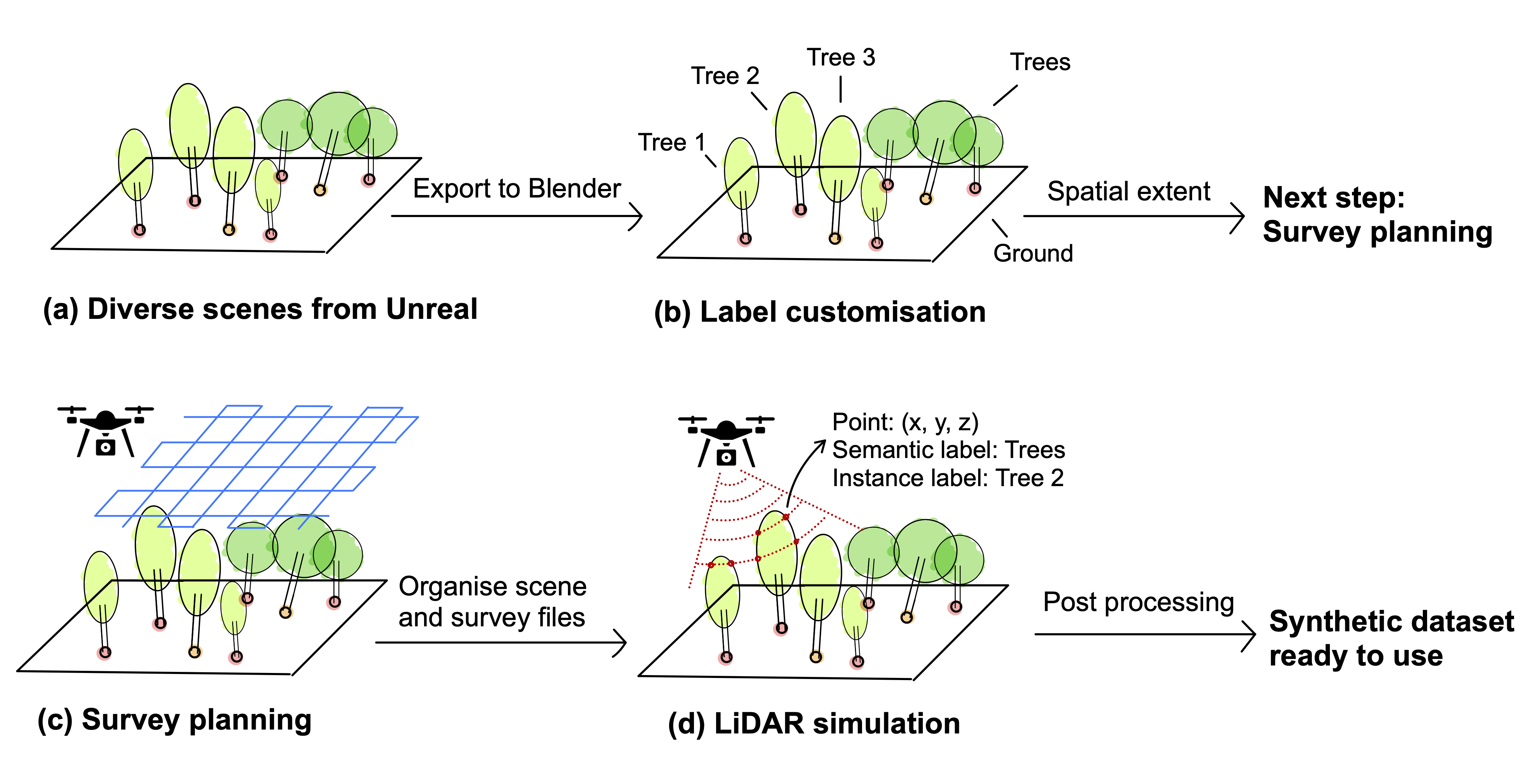}
    \caption{\textbf{Synthetic data generation pipeline, written in Python, and fully automated to execute each step in this figure.} \textit{The pipeline processes a forest scene from Unreal Engine by (a) adding files to Blender, (b) customizing instance and semantic labels, (c) planning the UAV flight path, setting up a virtual laser scanner, and (d) structuring files for LiDAR survey. It completes post-processing to produce a machine learning dataset. This process can be executed with a single command line, allowing parameter adjustments through arguments.}}
    \label{fig:methods_synthetic_data_generation_pipeline}
\end{figure}

\textbf{Virtual LiDAR survey.} Once we have obtained the forest scenes, we use the HELIOS simulator \citep{winiwarter2022virtual} to simulate UAV laser scans and generate point clouds with the desired annotations. To this end, we have developed a pipeline written in Python that will convert UE forest scenes, customise the labels, plan the virtual UAV flight for the survey and conduct the LiDAR simulation (\cref{fig:methods_synthetic_data_generation_pipeline}). 


An attractive aspect of synthetic data is that it can provide instance-level labels along with the data. 
Finer annotations, such as trunk, branch, or detailed foliage labels, are also possible but require prior customization of tree assets with specific material properties. These properties are not always present in the existing Unreal Engine assets, which are primarily developed for gaming purposes.
HELIOS generates labels by voxelizing the 3D objects for simulation. Each voxel carries the attributes specified in the 3D objects. During simulation, it will cast the ray through the voxel---their intersection will be recorded as the point location, and attributes of the voxel will be recorded as labels of the points. To enable the simulation of instance ID, we export the individual trees from Unreal Engine to Blender and then assign an unique ID as attribute for each object. A Blender add-on developed by \citet{neumann2022semantic} is integrated into our pipeline to convert forest scenes to a data structure that the simulator can read. 

Based on the customised forest, our pipeline will generate a survey file for each scene that will be used by HELIOS to conduct the survey. Our virtual survey resembles how UAV-laser scans work in the real world. \cref{tab:helios_ULS_survey_params} gives an overview of the key parameters of the survey. We use \texttt{RIEGL VUX-1 UAV}---a popular scanner model in real-world forest surveys---as our scanner. 
\texttt{Criss-cross} is used as the flight pattern, which, compared to the \texttt{Parallel} option, yields denser point clouds. 
The \texttt{Pulse Frequency} (100 kHz) and \texttt{Flight Speed} (5 m/s) jointly determine the sampling density along the track in our simulated survey. The chosen values match the typical RIEGL VUX-1 UAV survey settings in forest applications \citep{puliti2023instance}.
\texttt{Relative Altitude} is set at 60.0 m, which will work for all forest scenes except the RedWood scene, where it is set at 120 m given the size of the large tree. 

With the forest scenes and survey files ready, the pipeline will proceed with the LiDAR simulation to generate UAV-laser scans with the desired annotations. 

\begin{table}[htbp]
\centering
\caption{\textbf{Our virtual survey resembles a real world UAV mission.} \textit{The RIEGL VUX-1 UAV scanner is widely used. For \texttt{Redwood} scene we set the Relative Altitude as $120m$ because of the tree's large size. All parameters are customizable in our data pipeline.}}
\resizebox{0.5\textwidth}{!}{
\begin{tabular}{lc}
\toprule
\textbf{Parameter} & \textbf{Value} \\
\midrule
Scanner Model         & RIEGL VUX-1 UAV \\
Pulse Frequency       & 100 kHz \\
Max Returns (100 kHz) & 15 \\
Flight Pattern        & Criss-cross \\
Flight Spacing         & 20.0 m \\
Flight Speed          & 5.0 m/s \\
Relative Altitude     & 60.0 m \\
\bottomrule
\end{tabular}
}
\label{tab:helios_ULS_survey_params}
\end{table}

\subsection{Generated synthetic dataset}
\begin{figure}[htbp]
    \centering
    \includegraphics[width=\textwidth]{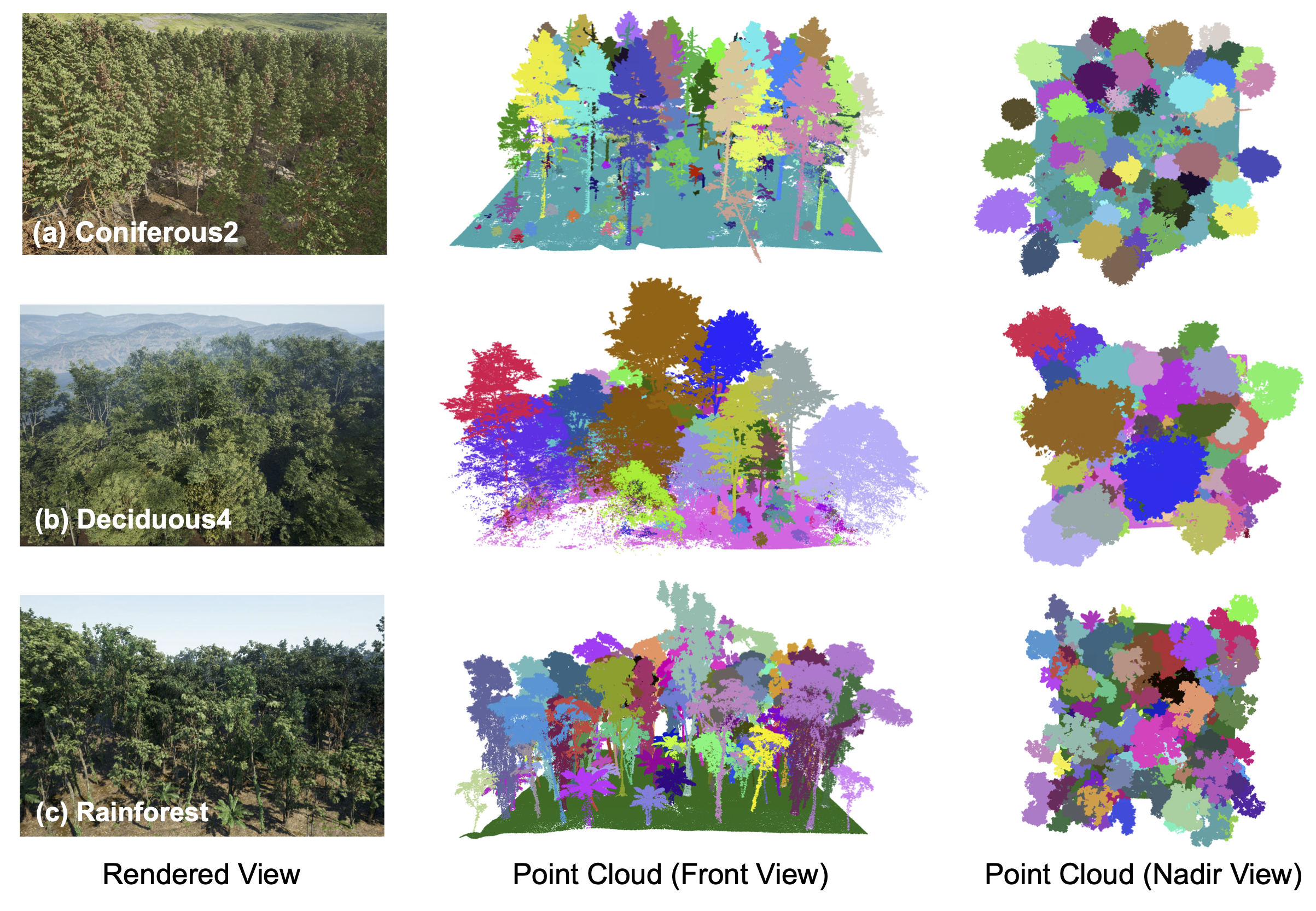}
    \caption{\textbf{The synthetic data is diverse, physics-based, and available at large scale.} \textit{Using Unreal Engine, large-scale, photorealistic forest scenes are created using gaming assets and procedural foliage. Then the pipeline simulates laser scans to produce synthetic point clouds with instance labels (shown here in colours). The 50m x 50m tiled plots are ready for machine learning.}}
    \label{fig:methods_synthetic_data_overview}
\end{figure}

The synthetic data generation pipeline simulates UAV laser scans for 12 forest scenes --- 5 adapted from the original SPREAD collection, and 7 new scenes (\cref{tab:spread_helios_data_overview_12types}). The dataset includes 4 coniferous and 6 deciduous scenes, focusing on European forests (\cref{fig:methods_synthetic_data_overview}). 
Two additional scenes, Rainforest and Redwood, are retained from the original SPREAD collection for completeness, but are not our main regional focus. 
The data set is prepared for machine learning by merging point clouds from each scene, tiling them into 50m x 50m plots, and splitting them into training (70\%), validation (15\%) and test sets (15\%). All plots have a point density greater than 1000 $\textrm{pts/m}^2$, adequate for effective tree segmentation learning \citep{wielgosz2024segmentanytree}.

\label{sec:methods}
\begin{table}[htbp]
\centering
\caption{\textbf{Synthetic data covers 12 forest scenes for a wide range of forest types, totalling 75 hectares.} \textit{Forest scenes in the first block are adapted from the SPREAD dataset \citep{feng2025spread}. The generated data is tiled into roughly 50m*50m plots and split into train, test, and validation sets.}}
\resizebox{\textwidth}{!}{
\begin{tabular}{l l ccc c}
\toprule
\textbf{Scene Name} & \textbf{Tree Species} & \multicolumn{3}{c}{\textbf{Plots}} & \textbf{points/m²} \\
\cmidrule(lr){3-5}
& & \textbf{Train} & \textbf{Val} & \textbf{Test} & \\
\midrule
Deciduous1 & Birch & 17 & 4 & 4 & 2092.687 \\
Deciduous2   & Beech, Oak & 17 & 4 & 4 & 1414.716 \\
Deciduous3     & Ash, Linden & 17 & 4 & 4 & 1638.955 \\
Rainforest  & Alii Fig, Palm, Fern, Rubber Fig, Umbrella & 17 & 4 & 4 & 1593.875 \\
Redwood     & Sequoia & 17 & 4 & 4 & 1120.249 \\
\midrule
Coniferous1      & Scots Pine, Silver Birch & 17 & 4 & 4 & 1693.415 \\
Coniferous2     & Scots Pine, Mountain Ash & 17 & 4 & 4 & 1576.841 \\
Coniferous3    & Fir, Spruce & 17 & 4 & 4 & 1944.650 \\
Coniferous4   & Scots Pine, Hornbeam & 17 & 4 & 4 & 1679.339 \\
Deciduous4    & Beech, Hazel, Norway Maple, Ash & 17 & 4 & 4 & 1604.083 \\
Deciduous5    & Black Alder, Hazel, Hornbeam, Ash & 17 & 4 & 4 & 1716.359 \\
Deciduous6   & Hornbeam, Hazel, Linden & 17 & 4 & 4 & 1767.841 \\
\bottomrule
\end{tabular}
}
\label{tab:spread_helios_data_overview_12types}
\end{table}

\subsection{Benchmark dataset from real world}
Models trained on our synthetic dataset needs to be validated on the real-world dataset to study the effectiveness of our approach. 
FOR-Instance~\citep{puliti2023instance} is a recently published UAV laser scan dataset that has previously been used for tree segmentation development and evaluation \citep{straker2023instance,xiang2024automated,wielgosz2024segmentanytree}, making it a suitable choice for this study.

\begin{figure}[htbp]
    \centering
    \includegraphics[width=\textwidth]{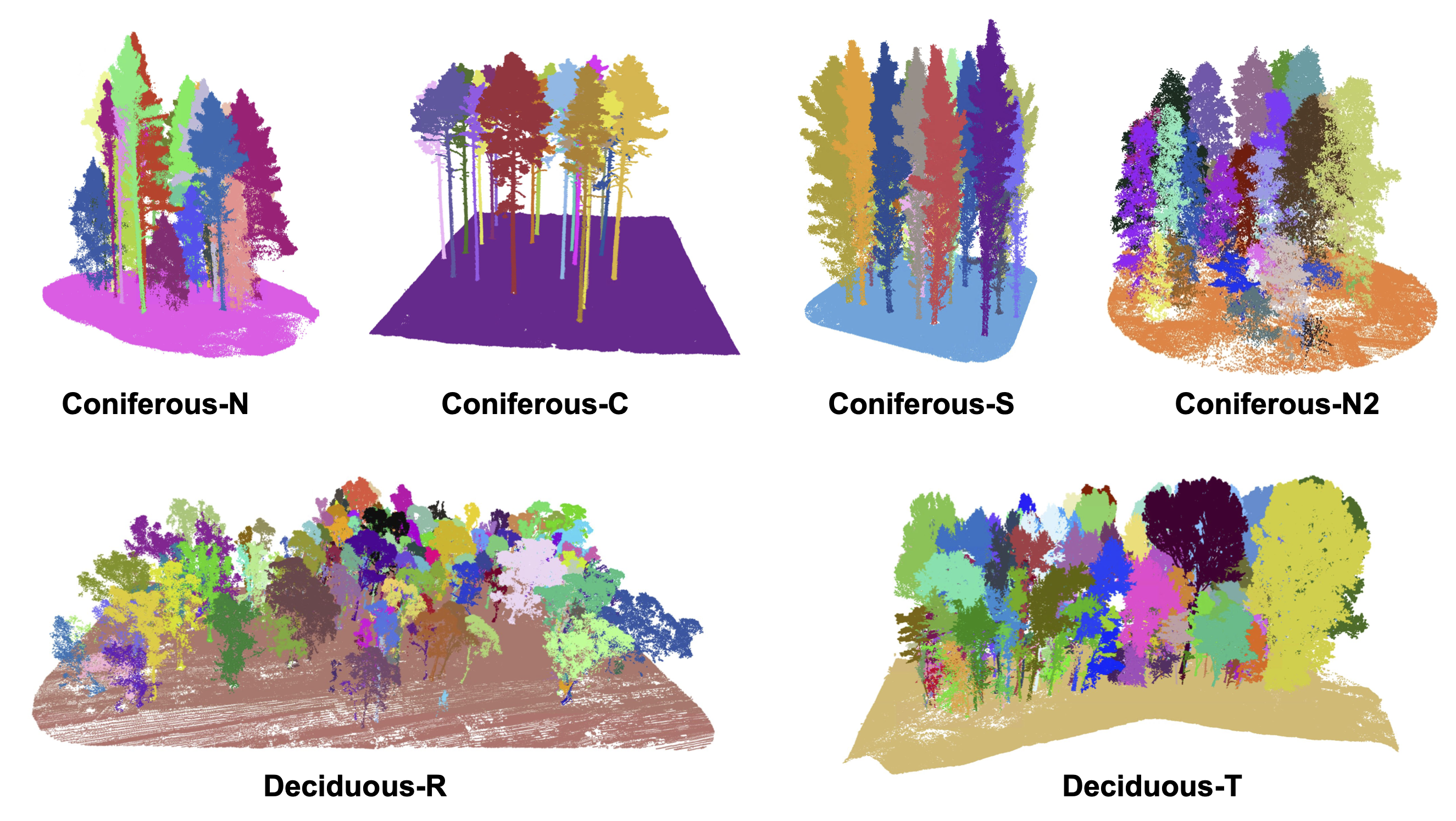}
    \caption{\textbf{Annotating the FOR-Instance dataset \citep{puliti2023instance} took two skilled annotators six months}, in addition to the time spent on data collection. \textit{It covers 3 coniferous scenes and 2 deciduous scenes totalling 2.79 hectares, and Coniferous-N2 is a supplementary dataset from\citet{xiang2024automated}. Annotated plots from each region are illustrated.}}
    \label{fig:methods_real_data_overview}
\end{figure}

The FOR-Instance data set covers 5 forest regions (\cref{tab:forinstance_overview_v2}), including 3 coniferous regions (Coniferous-N, Coniferous-S, Coniferous-S) and 2 deciduous regions (Deciduous-R, Deciduous-T), totalling 2.79 hectares of forest plots (\cref{fig:methods_real_data_overview}). These forest plots are collected mainly from Norway, Central Europe, Australia, and New Zealand. However, most of the forest plots consist of coniferous pine trees from the Coniferous-N region, while the rest of the region consists only of 1 or 2 forest plots. Scanning and annotating forest plots is costly, particularly for the complex deciduous forests.

\begin{table}[htbp]
\centering
\caption{\textbf{Overview of the FOR-Instance data set} from \citep{puliti2023instance}. \textit{NIBIO2 is not part of the FOR-Instance dataset but is included in the training set in \citep{xiang2024automated}. Note that RMIT and TUWIEN are deciduous forests and the rest are coniferous forests. Except for NIBIO and NIBIO2, the rest of the regions only have 1 or 2 training plots available.}}
\resizebox{\textwidth}{!}{
\begin{tabular}{l l l ccc c}
\toprule
\textbf{Scene Name} & \textbf{Original Name} & \textbf{Tree Species} & \multicolumn{3}{c}{\textbf{Plots}} & \textbf{Point Density (pts/m²)} \\
\cmidrule(lr){3-5}
& & & \textbf{Train} & \textbf{Val} & \textbf{Test} & \\
\midrule
Coniferous-N & NIBIO & Norway Spruce (dominated) & 8 & 6 & 6 & 9529 \\
Coniferous-C & CULS & Scots Pine & 1 & 1 & 1 & 2585 \\
Coniferous-S & SCION & Monterey Pine & 2 & 1 & 2 & 4576 \\
Deciduous-R & RMIT & Eucalyptus pulchella & 1 & 0 & 1 & 498 \\
Deciduous-T & TUWIEN & Deciduous species in Austria & 1 & 0 & 1 & 1717 \\
\midrule
Coniferous-N2 & NIBIO2 & Norway Spruce (dominated) & 29 & 6 & 15 & $> 1000$ \\
\bottomrule
\end{tabular}
}
\label{tab:forinstance_overview_v2}
\end{table}

FOR-Instance contains both instance and semantic segmentation labels. Each tree in the point clouds has a unique  segmentation ID, similar to our synthetic dataset.
Semantic labels cover five classes: stem, woody branches, live branches, low-vegetation, and ground. Our synthetic dataset currently offers only binary labels (tree and non-tree) --- generating the full 5-class labels would require customising 3D tree assets with distinct material properties, which is not readily available in the high-quality gaming assets we use.
For consistency therefore, we remapped the original 5 semantic labels of the FOR-Instance dataset into the binary classes (tree: woody branches, live branches, low-vegetation. non-tree: low-vegetation, ground). 
Our synthetic data collection method produces forest plots with point densities exceeding 1000 $pts/m^2$, consistent with those in the FOR-Instance data set collected by UAV laser scanning.

\subsection{Tree segmentation algorithm}\label{sec:tree_segmentation_algorithm}
\begin{figure}[htbp]
    \centering
    \includegraphics[width=\textwidth]{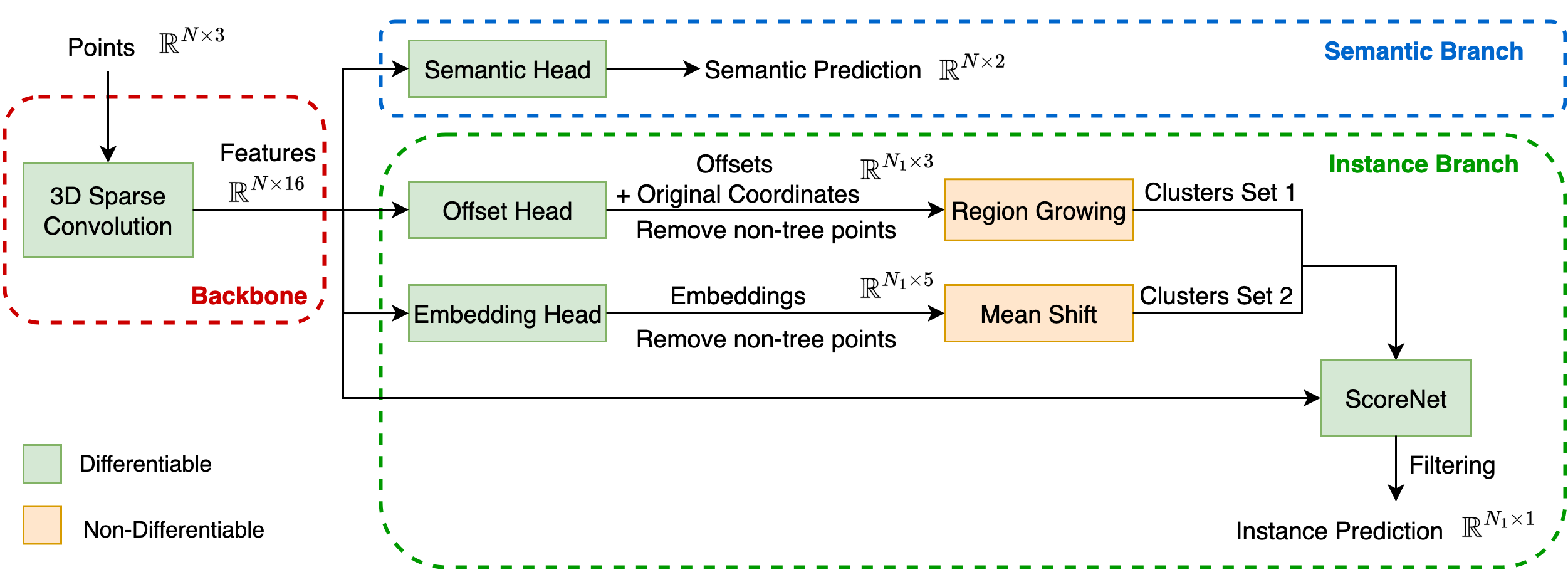}
    \caption{\textbf{The panoptic tree segmentation algorithm~\citep{xiang2024automated} $\mathcal{A}$ adapted for this study.} \textit{Panoptic segmentation addresses instance and semantic segmentation simultaneously. $\mathcal{A}$ uses the Minkowski Engine, a 3D CNN, as its backbone followed by semantic and instance branches. In our experiments, we simplify semantic prediction from five classes to binary: tree and non-tree.}}
    \label{fig:methods_xiang_tree_segmentation_algorithm}
\end{figure}

For the tree segmentation algorithm, we use ForAINet from \citet{xiang2024automated} (denoted by $\mathcal{A}$), which was developed using the FOR-Instance dataset and considered the state-of-the-art algorithm for instance and semantic segmentation of forest point clouds (\cref{fig:methods_xiang_tree_segmentation_algorithm}). 

In summary, $\mathcal{A}$ is adapted from the PointGroup algorithm \citep{jiang2020pointgroup} and uses Minkowski Engine \citep{choy20194d}, a U-Net for sparse 3D convolutions in point clouds,  as its backbone. The backbone is followed by three heads. 
The semantic head takes backbone features at each point and classifies them into the five semantic classes. 
Two other heads that are complementary for the instance segmentation: 
\begin{enumerate}
    \item an embedding head that embeds the backbone feature into a 5 dimensional space and then performs a clustering with mean-shift in this 5D space to detect tree clusters from embeddings;
    \item an offset head that shifts the points in the original 3D space via predicted 3D offsets, followed by another clustering with region-growing to detect tree clusters. 
\end{enumerate}

Finally, the two instance segmentation heads are followed by a small neural network module ScoreNet \citep{jiang2020pointgroup} that learns to score each detected cluster by comparing ground truth IoU with predicted IoU. During inference, these scores, together with the detected clusters, are passed to Non-Maximal Suppression to obtain the final predictions of instance segmentation.

To train the model, 8m radius cylinders are randomly sampled from the forest plots as individual data samples. In addition to the basic training setting, \citet{xiang2024automated} applied a data augmentation strategy called `TreeMix3D', which they found to boost model performance. This augmentation increases forest structure diversity by randomly replacing 30\% of the trees from one cylinder sample with trees from another cylinder sample. 
During inference, the model use grid cylinder sampling, placing cylinders systematically to cover the plot; each is inferred independently, with results optionally merged for plot-level inference.
The 8m-radius cylinders are mainly designed for tree sizes commonly seen in deciduous and coniferous plots. For consistency, we exclude the RedWood scene from model training, since the large tree sizes makes them unsuitable for this segmentation algorithm. 

Our experiments studying the effectiveness of synthetic data, use a simplified form $\mathcal{A}_{\text{simp}}$ of the original algorithm which we now denote $\mathcal{A}_{\text{full}}$:
\begin{itemize}
    \item The semantic head uses just 2 binary classes, as already explained. 
    \item Instead of jointly training ScoreNet with other modules from very early epochs, we first train modules and then save the detected clusters to train ScoreNet separately. Costly non-differentiable clustering is then done just once, after other modules have been fully trained, significantly speeding up training on large-scale synthetic data. 
\end{itemize}

\subsection{Experiments}
We first pre-train $\mathcal{A}_{\text{simp}}$ on the full-scale synthetic Sim\textsubscript{11,187} data for 150 epochs, where `Sim' denotes synthetic data, `Real' denotes real FOR-Instance data, the first subscript indicates the number of scenes, and the second the number of plots. 
Sim\textsubscript{11,187} consists of 11 scenes, totalling 13635 cylinder samples (see also \cref{tab:experimental_settings_syn_ablations} for scene compositions). This pre-training should provide a strong feature extraction network benefiting from the diversity of the synthetic data that includes coniferous, deciduous, and rainforest scenes. The pretrained model is then fine-tuned on the corresponding coniferous and deciduous forest plots from the real FOR-Instance dataset. 
Unless otherwise specified, we report metrics as the mean over all cylinder samples obtained during inference, each with its own accuracy, recall, and F-score. 

We have built two few shot datasets, subsets of FOR-Instance with few labelled plots to assess model performance with limited data:
\begin{itemize}
    \item Real-C\textsubscript{3,3}, contains the 3 coniferous scenes from the original FOR-Instance dataset, one plot from each coniferous region (Real-C-N\textsubscript{1,1}, Real-C-C\textsubscript{1,1}, and Real-C-S\textsubscript{1,1}).
    \item Real-D\textsubscript{2,2}, contains the 2 deciduous scenes from FOR-Instance dataset, one plot from each deciduous region (Real-D-R\textsubscript{1,1} and Real-D-T\textsubscript{1,1}). (These are the only deciduous plots available in the FOR-Instance dataset due to the difficulty in scanning and annotating deciduous forests.) 
\end{itemize}

Real-C\textsubscript{3,3} only accounts for 7.2\% of the total available samples, and for Real-D\textsubscript{2,2} it is 15.2\% (\cref{tab:experimental_settings_for}). We fine-tune pretrained models on each few-shot dataset and compare them to models trained from scratch. This comparison will show how much $\mathcal{A}_{\text{simp}}$ benefits from pre-training on synthetic data versus training solely on limited real data, crucial for effective few-shot learning.
We also try fine-tuning the pretrained model on a single real plot of each individual forest region from the few-shot datasets. This is an even more extreme case of few-shot. It is also a practical scenario for a forest manager with only a single annotated forest plot available, and happens in the FOR-Instance dataset, where most regions, except Coniferous-N and Coniferous-N2, have just 1 or 2 plots.

We also compare our few-shot learning results with the results obtained from training with $\mathcal{A}_{\text{simp}}$ on the full-scale FOR-Instance dataset Real\textsubscript{6,42}, which contains 6 scenes, totalling 42 plots and 1051 samples (\cref{tab:experimental_settings_for}), to see if we can obtain competitive results with only a few forest plots. 
\begin{table}[htbp!]
\centering
\caption{\textbf{Real datasets used in  few shot learning experiments.}  \textit{The subscripts denote the number of scenes and the number of plots. Besides the full dataset Real\textsubscript{6,42}, we organised few-shot subsets for coniferous (Real-C) and deciduous forests (Real-D). We  report cylinder sample counts from each subset  and their percentages of the full dataset.}}
\resizebox{\textwidth}{!}{
\begin{tabular}{l l c c c}
\toprule
\textbf{Dataset} & \textbf{Scenes} & \textbf{Plots} & \textbf{Samples} & \textbf{Percentage (\%)} \\
\midrule
Real\textsubscript{6,42} & Coniferous-N, Coniferous-C, Coniferous-S, Coniferous-N2 & \multirow{2}{*}{42} & \multirow{2}{*}{1051} & \multirow{2}{*}{100} \\
& Deciduous-R, Deciduous-T \\
\midrule
Real-C\textsubscript{3,3} & Coniferous-N, Coniferous-C, Coniferous-S & 3 & 76 & 7.2 \\
Real-C-N\textsubscript{1,1} & Coniferous-N & 1 & 32 & 3.1 \\
Real-C-C\textsubscript{1,1} & Coniferous-C & 1 & 31 & 2.9 \\
Real-C-S\textsubscript{1,1} & Coniferous-S & 1 & 23 & 2.2 \\
\midrule
Real-D\textsubscript{2,2} & Deciduous-R, Deciduous-T & 2 & 160 & 15.2 \\
Real-D-R\textsubscript{1,1} & Deciduous-R & 1 & 72 & 6.8 \\
Real-D-T\textsubscript{1,1} & Deciduous-T & 1 & 88 & 8.4 \\
\bottomrule
\end{tabular}
}
\label{tab:experimental_settings_for}
\end{table}

\subsection{Implementation Details}\label{sec:impl_details}
Pretraining on the synthetic dataset Sim\textsubscript{11,187} totals 150 epochs. For fine-tuning, the pretrained models are fine-tuned on real datasets for 60 epochs. Baselines are built by training the models from scratch on the real datasets for 120 epochs. 
All models are trained with the data augmentation setting `TreeMix3D' as discussed in \cref{sec:tree_segmentation_algorithm}, which is in line with \citep{xiang2024automated}. 
ScoreNet is not initially included in training because the point clustering required would substantially slow down the process. Instead, once other layers are trained, detected clusters are saved and used to train ScoreNet separately (for 30 epochs) to score and filter the clusters. The initial learning rate for pretraining is 0.001, and the initial learning rate for fine-tuning is 0.0001. The batch size is 8. 
All models are trained on a NVIDIA GeForce RTX 4080 GPU, with 128G RAM. 

%% file: 4_results.tex
\section{Results} \label{sec:results}
\subsection{Synthetic data can enhance instance segmentation with minimal use of real plots}\label{sec:few_shot_learning_boost}
In \cref{tab:eval_metrics_fine_tune_coni_n_deci_r}, we compare the performance of $\mathcal{A}_{\text{simp}}$ trained in different settings in coniferous and deciduous plots. More detail can be found in \cref{tab:eval_metrics_fine_tune_coniferous_appx} and \cref{tab:eval_metrics_fine_tune_deciduous_appx} in \cref{sec:few_shot_learning_boost_appx}.

\begin{table}[htbp!]
\centering
\caption{\textbf{Pretraining on synthetic data,  and using minimal real data fine-tuning, performs competitively with a much larger, labelled, real dataset.} \textit{F1 scores improve significantly for small, labelled coniferous regions when synthetic (mixed deciduous/coniferous) data is available for pre-training. Deciduous plots show a similar pattern, with the same pretraining data.}}
\label{tab:eval_metrics_fine_tune_coni_n_deci_r}
\resizebox{\textwidth}{!}{
\begin{tabular}{llllcc}
\toprule
\textbf{Test} & \textbf{Pretrain} & \textbf{Train/Tune} & \textbf{Experiment} & \textbf{Real Data Used (\%)} & \textbf{F1-score (\%)} \\ \midrule

\multirow{4}{*}{Coniferous-N}  & - & Real\textsubscript{6,42} & Training on all real plots from scratch & 100 & 79.8 \\
                        & - & Real-C\textsubscript{3,3} & Training on 3 coniferous plots from scratch & 7.2 & 50.5 \\
                        & Sim\textsubscript{11,187} & Real-C\textsubscript{3,3} & Fine-tuning on 3 coniferous plots & 7.2 & 79.4 \\
                        & Sim\textsubscript{11,187} & Real-C-N\textsubscript{1,1} & Fine-tuning on 1 plot from Coniferous-N & 3.1 & 77.1 \\ \midrule
\multirow{4}{*}{Deciduous-R}  & - & Real\textsubscript{6,42} & Training on all real plots from scratch & 100 & 69.9 \\
                        & - & Real-D\textsubscript{2,2} & Training on 2 deciduous plots from scratch & 15.2 & 31.6 \\
                        & Sim\textsubscript{11,187} & Real-D\textsubscript{2,2} & Fine-tuning on 2 deciduous plots & 15.2 & 70.0 \\
                        & Sim\textsubscript{11,187} & Real-D-R\textsubscript{1,1} & Fine-tuning on 1 plot from Deciduous-R & 6.8 & 69.3 \\
\bottomrule
\end{tabular}
}
\end{table}


\begin{figure}[htbp]
    \centering
    \includegraphics[width=\textwidth]{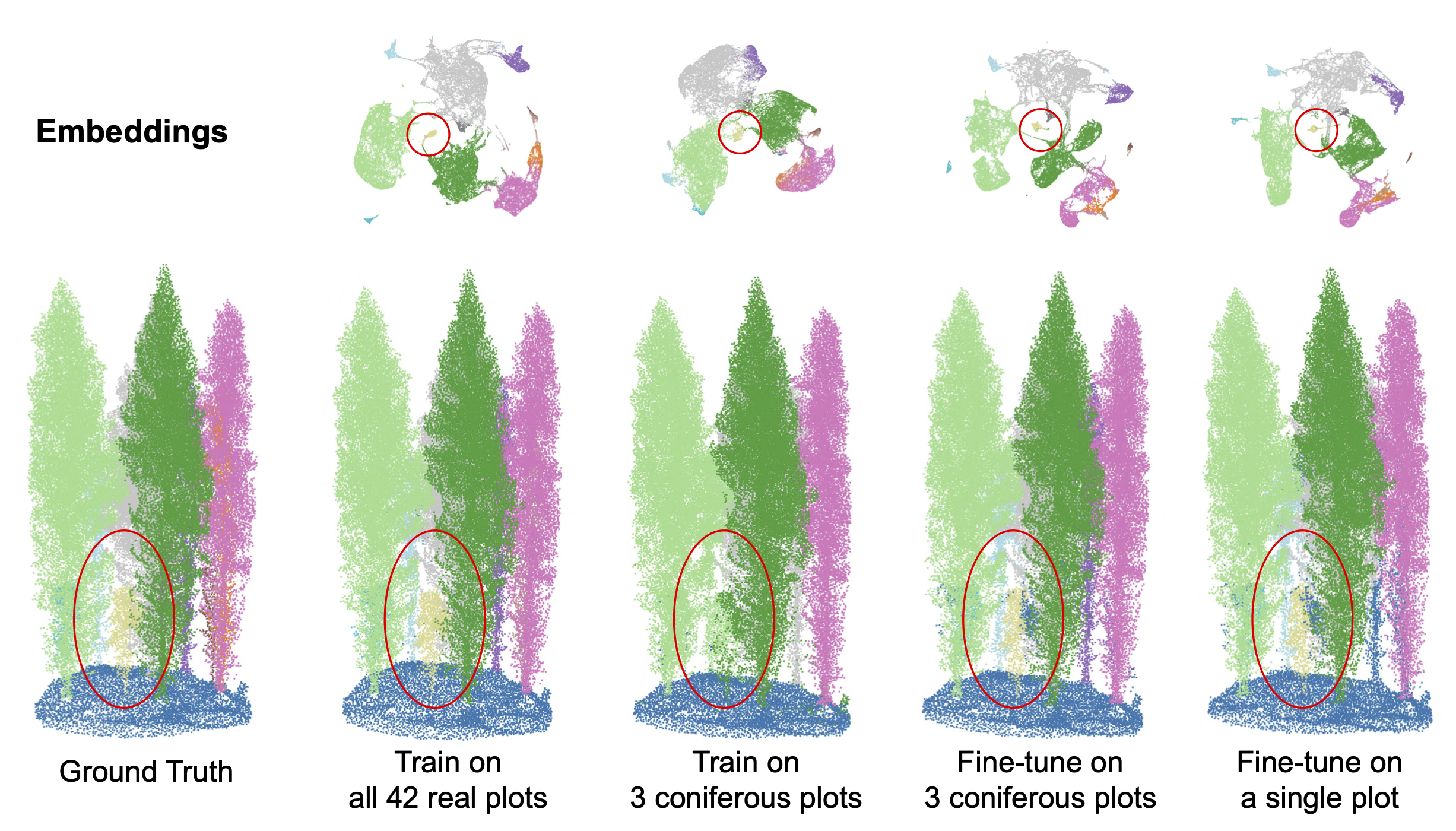}
    \caption{\textbf{Pretraining on synthetic data enables the algorithm to learn representations like those from real data, with fine-tuning on just a single forest plot.} \textit{We present a sample from Coniferous-N, where embeddings and points shifted by offsets serve as the foundation, for instance segmentation. Both embeddings and points are colourized with ground-truth labels and projected in 2D using UMAP \citep{mcinnes2018umap}. Training only on three coniferous plots, the algorithm struggles to learn disentangled representations, as seen in segmentation results where small trees in the understory are not segmented, noted within the red circle.}}
    \label{fig:results_few_shot_segmentation}
\end{figure}

Training from scratch with the reduced coniferous data set Real-C\textsubscript{3,3} leads to a significant drop in F1 score of more than 29\%. Pretraining with synthetic data and then fine-tuning with Real-C\textsubscript{3,3} fully restores the drop in performance. (In fact this improvement is evident in all coniferous regions: see \cref{tab:eval_metrics_fine_tune_coniferous_appx}.) Even extreme few-shot training, using only 3.1\% of the samples, is competitive, dropping just a couple of percentage points on F1. 
\cref{fig:results_few_shot_segmentation} shows learnt representations and segmentation results for a Coniferous-N sample in various settings; synthetic data pretraining facilitates the learning of comparable representations, even with minimal fine-tuning.

A similar pattern of improvement can be observed in Deciduous-R (\cref{tab:eval_metrics_fine_tune_coni_n_deci_r}) and other deciduous plots (\cref{tab:eval_metrics_fine_tune_deciduous_appx}). 
Training only on Real-D\textsubscript{2,2} reduces the F1 score significantly compared to the full-scale Real\textsubscript{6,42}. For Deciduous-R, it drops from 69.9\% to 31.6\%, despite Real-D\textsubscript{2,2} covering all available plots. Fine-tuning the pretrained model on Real-D\textsubscript{2,2} fully restores the loss.

In terms of computational efficiency, fine-tuning the model needs only 60 epochs, compared to 120 epochs when training from scratch. 
Our synthetic data approach to pre-training achieves competitive segmentation performance, with substantial savings both of annotation effort and computational cost. 

\subsection{Key elements contributing to the effectiveness of synthetic data}\label{sec:ablation_key_factors}
In \cref{sec:intro}, we point out three key strengths of our synthetic data generation pipeline: diverse forest scenes, large-scale forest generation, and physics-based laser scans for point generation. 
We perform ablation studies to study the role of each of the three (\cref{tab:experimental_settings_syn_ablations}). 
We pre-train $\mathcal{A}_{\text{simp}}$ on each of these synthetic datasets in the same setting of pre-training on Sim\textsubscript{11,187}, as described in \cref{sec:impl_details}. 
Then the zero-shot learning experiments  directly evaluate the model obtained in each forest region from the real datasets. 

\begin{table}[htbp!]
\centering
\caption{\textbf{Synthetic datasets in our ablation study analyze three factors:} simulation physics (Nodal\textsubscript{4,20} vs. Sim\textsubscript{4,20}), scene diversity (Sim\textsubscript{4,20} vs. Sim\textsubscript{11,55}), and dataset scale (Sim\textsubscript{11,55} vs. Sim\textsubscript{11,187}).\textit{Nodal points derive from tree mesh nodes, while Sim points come from LiDAR simulations. Both Sim\textsubscript{4,20} and Sim\textsubscript{11,55} cover 30\% of available plots (5 plots per scene) from their scenes.}}
\resizebox{\textwidth}{!}{
\begin{tabular}{l l c c}
\toprule
\textbf{Dataset} & \textbf{Scenes} & \textbf{Plots} & \textbf{LiDAR}  \\
\midrule
Nodal\textsubscript{4,20} & Deciduous1, Deciduous2, Deciduous3, Rainforest & 20 & \xmark \\
Sim\textsubscript{4,20} & Deciduous1, Deciduous2, Deciduous3, Rainforest & 20 & \cmark \\
\multirow{2}{*}{Sim\textsubscript{11,55}} & Deciduous1, Deciduous2, Deciduous3, Deciduous4, Deciduous5, Deciduous6, &  \multirow{2}{*}{55} & \multirow{2}{*}{\cmark} \\
& Coniferous1, Coniferous2, Coniferous3, Coniferous4, Rainforest \\
\multirow{2}{*}{Sim\textsubscript{11,187}} & Deciduous1, Deciduous2, Deciduous3, Deciduous4, Deciduous5, Deciduous6, &  \multirow{2}{*}{187} & \multirow{2}{*}{\cmark} \\
& Coniferous1, Coniferous2, Coniferous3, Coniferous4, Rainforest \\
\bottomrule
\end{tabular}
}
\label{tab:experimental_settings_syn_ablations}
\end{table}

\begin{figure}[htbp]
    \centering
    \includegraphics[width=\textwidth]{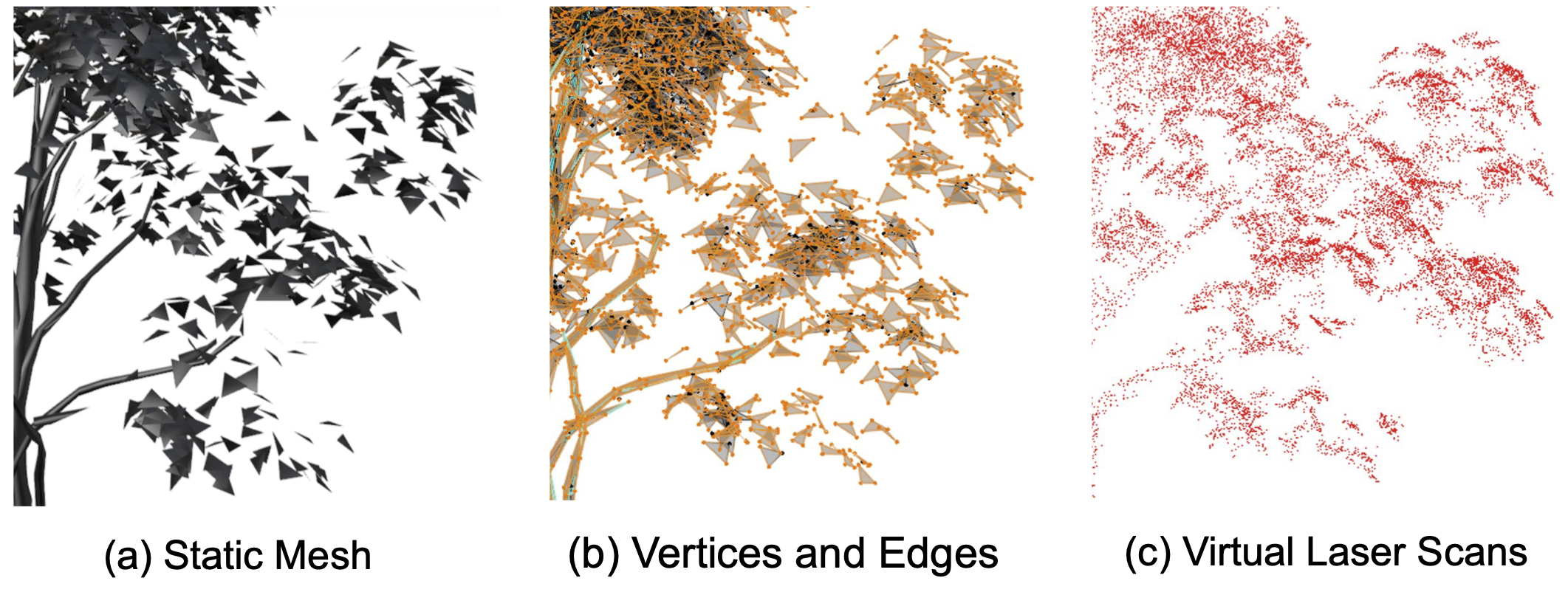}
    \caption{\textbf{Nodal points from tree meshes provide a simple way to obtain point clouds.} \textit{A large ash tree model from the Deciduous3 scene is shown. A rendered tree model consists of textures and a static mesh. The static mesh, made of vertices and edges, directly yields nodal points as point clouds. LiDAR simulation point clouds are based on UAV laser scans and are denser than these nodal points.}}
    \label{fig:results_nodal_vs_lidar_points}
\end{figure}

We use four forest scenes from the SPREAD dataset \citep{feng2025spread}, each with 5 plots, totalling 20 plots. The data set incorporates tree mesh points that arguably approximate point clouds (\cref{fig:results_nodal_vs_lidar_points}). We refer to the point clouds from tree mesh points as Nodal\textsubscript{4,20}, and those from UAV-LiDAR simulation with HELIOS as Sim\textsubscript{4,20} (\cref{tab:experimental_settings_syn_ablations}).

\cref{tab:eval_metrics_pretrain_ablations} shows the results for Coniferous-C. More detail can be found in \cref{tab:eval_metrics_pretrain_ablations_appx} in \cref{sec:ablation_key_factors_appx}. Training with UAV-LiDAR simulation (Sim\textsubscript{4,20}) significantly enhances performance, whereas nodal point clouds from tree mesh (Nodal\textsubscript{4,20}) performs poorly: F1 score in \cref{tab:eval_metrics_pretrain_ablations} improved from 5.4\% to 42.6\%. A similar pattern of improvement can be observed for other regions (\cref{tab:eval_metrics_pretrain_ablations_appx} in \cref{sec:ablation_key_factors_appx}).
Note that HELIOS ray-tracing simulates UAV scanning, producing over 1000 points/m², unlike tree meshes, which average 100-200 points/m² and have different distributions with denser canopy points from laser scans.

\begin{table}[htbp!]
\centering
\caption{\textbf{Three factors enhance the value of synthetic data for training: physics-based laser scans, scene diversity, and increased data.} \textit{Adding physics-based LiDAR simulation (Sim\textsubscript{4,20}) improves F1 score over using tree mesh nodal points (Nodal\textsubscript{4,20}). Greater scene diversity further boosts performance (Sim\textsubscript{11,55} versus Sim\textsubscript{4,20}), with additional data providing a smaller gain (Sim\textsubscript{11,187} versus Sim\textsubscript{11,55}).}}

\label{tab:eval_metrics_pretrain_ablations}
\resizebox{\textwidth}{!}{
\begin{tabular}{lllc}
\toprule
\textbf{Test} & \textbf{Train} & \textbf{Ablation Experiment} & \textbf{F1-score} (\%) \\ \midrule
\multirow{4}{*}{Coniferous-C}  
      & Nodal\textsubscript{4,20} & Nodal points from 4 scenes (5 plots per scene) & 5.4 \\
      & Sim\textsubscript{4,20} & Simulated LiDAR from the same 4 scenes (5 plots per scene) & 42.6 \\
      & Sim\textsubscript{11,55} & +7 additional scenes (total 11 scenes, 5 plots each) & 81.5 \\
      & Sim\textsubscript{11,187} & +More plots per scene (total 11 scenes, 17 plots each) & 87.0 \\
\bottomrule
\end{tabular}
}
\end{table}

We extend Sim\textsubscript{4,20} to Sim\textsubscript{11,55} to study scene diversity. Sim\textsubscript{11,20} adds three deciduous and four coniferous scenes not in Sim\textsubscript{4,20} and Sim\textsubscript{4,20}, with each scene having 5 plots like Sim\textsubscript{4,20}. There is consistent improvement across all regions (\cref{tab:eval_metrics_pretrain_ablations_appx}), with Coniferous-C's F-1 score increasing by 38.9\% from 42.6\% to 81.5\%.

We use the full-scale synthetic data set Sim\textsubscript{11,187} to study the effects of the size of the data set. Compared with Sim\textsubscript{11,55}, it includes all available plots (17 per forest scene) for the 11 scenes. More training data helps the model learn scene variations and enhances F1 performance from 81.5\% to 87\%.

%
\begin{figure}[htbp]
    \centering
    \includegraphics[width=\textwidth]{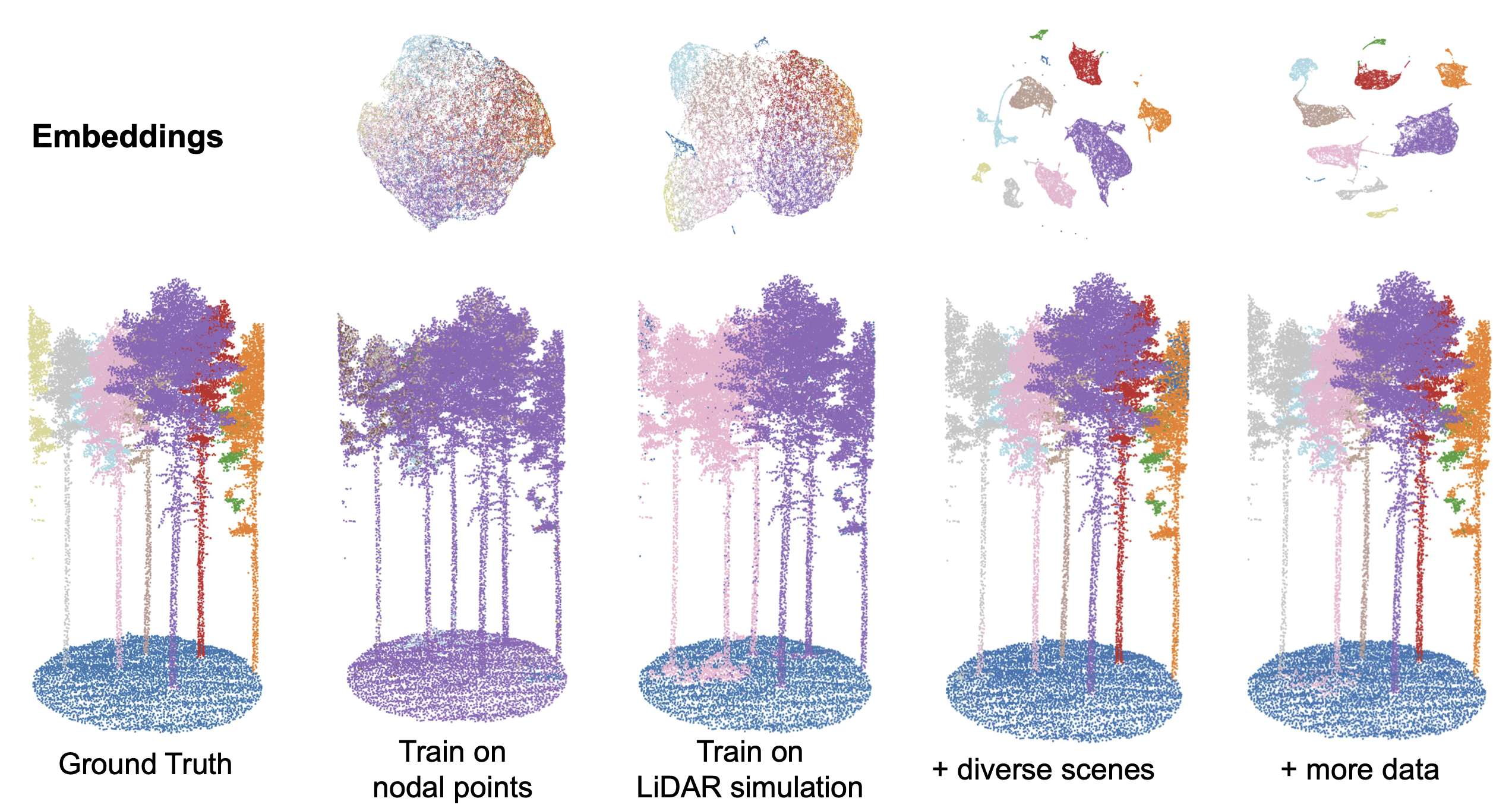}
    \caption{\textbf{Synthetic data helps the algorithm learn disentangled representations.} \textit{A Coniferous-C data sample with embeddings and offset-shifted points, is colorized with ground truth labels and projected into 2D using UMAP \citep{mcinnes2018umap}. Compared with training at nodal points, disentanglement is enhanced as we add: LiDAR simulation;  diverse scenes, especially newly included coniferous ones; more simulated data.}}
    \label{fig:results_zero_shot_segmentation}
\end{figure}

To summarize, the algorithm effectively learns about forest structures from synthetic data, crucial for real-world applicability --- see few-shot studies in (\cref{sec:few_shot_learning_boost}). Enhancing scene diversity seems to offer greater benefits than simply expanding the size of the dataset presumably by playing a key role in learning disentangled representations (\cref{fig:results_zero_shot_segmentation}).

\subsection{Bridging domain gaps with mixed training of synthetic and real data}\label{sec:mixed_training}
In \cref{sec:few_shot_learning_boost}, we conduct few-shot experiments with $\mathcal{A}_{\text{simp}}$ to show the effectiveness of the synthetic data. 
However, pretraining only on synthetic data may not produce representations as strong or generalisable as those learnt from a combination of synthetic and real data. 
Furthermore, the original real data includes five semantic labels for the model to learn, representing not only a practical application, but also an enriched multi-task learning scenario that could improve algorithm performance. 

We consider the synthetic data as the source domain ($\mathcal{S}$) and the real data as the target domain ($\mathcal{T}$). In practice, $\mathcal{T}$ can be unlabelled ($\mathcal{T}_{\text{unlabel}}$), as is common in forest surveys, or labelled ($\mathcal{T}_{\text{label}}$), as in the FOR-Instance dataset.
In this section, we focus on the labelled setting to examine whether pretraining in a combination of $\mathcal{S}$ and $\mathcal{T}_{\text{label}}$, including its full 5-class semantic labels, can produce stronger performance. To investigate this, we adopt the original algorithm $\mathcal{A}_{\text{full}}$ in a mixed training set. 

In mixed training, we use datasets from both $\mathcal{S}$ and $\mathcal{T}_{\text{label}}$. Although $\mathcal{A}_{\text{full}}$ is designed for five semantic classes, $\mathcal{S}$ provides only binary semantic labels (tree / non-tree). To accommodate both, the semantic branch is modified to have two heads: one predicts five classes for $\mathcal{T}_{\text{label}}$, and the other predicts binary labels for $\mathcal{S}$. The instance branch remains unchanged. Each batch contains an equal number of synthetic and real samples.
After pretraining, we continue to fine-tune the model on only the real data set from $\mathcal{T}_{\text{label}}$, improving alignment with the real data. 
We pretrain on the mixed dataset for 120 epochs and fine-tuned solely on the real dataset for 30 epochs. To accelerate training, $\mathcal{A}_{\text{full}}$ is initialized with the pretrained weights of Sim\textsubscript{11,187}, acquired in previous experiments. 

In \cref{tab:mixed_training_f1_comparison}, mixed-training F1 scores are broadly consistent with those reported by \citet{xiang2024automated} except for the  Deciduous-T plot,  where both our mixed training and our reproduction of their official model produce substantially lower results. This drop may result from random seed effects or the small size of the test set (35 trees), although the exact cause is uncertain. Overall, the comparison indicates that mixed training can match state-of-the-art performance but does not surpass it.

\begin{table}[htbp!]
\centering
\caption{\textbf{Our model trained on mixed data achieves results comparable to state-of-the-art, without exceeding it.} \textit{F1-scores for mixed training, results in \citet{xiang2024automated}, and our reproduction of \citet{xiang2024automated}, are similar across regions. Deciduous-T is an exception where both our result and the reproduction are notably lower than reported by \citet{xiang2024automated}. Metrics are computed at the plot level by aggregating predictions from all cylinder samples within each plot, following the same procedure of \citet{xiang2024automated} to ensure comparability.}}
\label{tab:mixed_training_f1_comparison}
\resizebox{\textwidth}{!}{
\begin{tabular}{lccc}
\toprule
\textbf{Test Set} & \textbf{Mixed Training} (\%) & \textbf{Reported in \citet{xiang2024automated}} (\%) & \textbf{Reproduced} (\%) \\ \midrule
Coniferous-N & 92.0 & 92.4 & 91.6 \\
Coniferous-C & 90.5 & 93.0 & 93.0 \\
Coniferous-S & 92.9 & 91.5 & 89.4 \\
Deciduous-R & 66.1 & 69.5 & 67.2 \\
Deciduous-T & 47.4 & 69.4 & 46.6 \\ \bottomrule
\end{tabular}
}
\end{table}

This performance on par is somewhat surprising, since incorporating more data should help the algorithm learn a stronger representation than training solely on $\mathcal{S}$ or $\mathcal{T}_{\text{label}}$, enhancing robustness and generalisation. 
A possible explanation may be that our test sets originate from the same forest plots as $\mathcal{T}_{\text{label}}$ reducing the scope for boost effects. 
In future, therefore it may be worthwhile to test models pretrained on mixed data from other regions, such as the Wytham Wood data set \citep{calders2022laser}. Alternatively, mixed training experiments could use a reduced set $\mathcal{T}_{\text{label}}$, excluding data from certain regions e.g. Coniferous-N2, to determine if similar state-of-the-art results can still be achieved with less real data. Further suggestions are provided in \cref{sec:advance_real_forest_vision}.

\section{Discussion}\label{sec:discussions}
\subsection{Improving the reusability of real-world labelled scans with procedural foliage generation}\label{sec:efficient_field_sampling}



High-quality field scans remain significant for our synthetic data pipeline, as they provide the detailed, labelled laser scans of individual trees needed to create realistic 3D tree assets. Traditionally, new study areas require scanning entire forest plots and manually annotating each tree to produce training data, a process that is time-consuming and costly \citep{calders2022laser,puliti2023instance,weiser2022individual}. 

The procedural foliage generation ($\mathcal{A}_{\text{PFG}}$) offers a more efficient alternative. Once a library of 3D tree assets has been created from existing field scans, $\mathcal{A}_{\text{PFG}}$ can recombine these assets to generate new forest plots customised to a given region of interest. This allows researchers to produce large, labelled datasets for new areas without rescanning or re-annotating whole plots, greatly improving the reusability of field-collected data.

For example, the scene `Deciduous3' in our synthetic dataset is a 250 m × 250 m (6.25 ha) forest containing 2,053 trees (328 trees/ha). These are generated from 17 scanned tree meshes (two species: Ash and Linden), scaled, rotated, and placed under procedural rules to represent different canopy levels. Even though this procedurally generated scene has limited species diversity, it can still capture the structural variability needed for effective model training, offering a practical balance between ecological realism and machine learning efficiency.

\begin{table}[htbp]
\centering
\caption{\textbf{A structurally diverse forest scene can be created using a small set of tree models.} \textit{The Deciduous3 composition uses 2,053 tree meshes within a 250m x 250m plot, derived from 17 basic meshes with various canopy levels, as shown in this table.}}
\resizebox{0.5\textwidth}{!}{
\begin{tabular}{lcc}
\toprule
\textbf{Mesh Name} & \textbf{Count} & \textbf{Percentage (\%)} \\
\midrule
TreesAshLarge\_A       & 60  & 2.92 \\
TreesAshLarge\_B       & 43  & 2.09 \\
TreesAshLarge\_C       & 42  & 2.05 \\
TreesAshMedium\_B      & 343 & 16.71 \\
TreesAshMedium\_C      & 180 & 8.77 \\
TreesAshSapling\_A     & 14  & 0.68 \\
TreesAshSapling\_B     & 32  & 1.56 \\
TreesAshSapling\_C     & 172 & 8.38 \\
TreesLindenLarge\_A    & 2   & 0.10 \\
TreesLindenLarge\_B    & 2   & 0.10 \\
TreesLindenLarge\_C    & 1   & 0.05 \\
TreesLindenMedium\_A   & 183 & 8.91 \\
TreesLindenMedium\_B   & 173 & 8.43 \\
TreesLindenMedium\_C   & 136 & 6.62 \\
TreesLindenSapling\_A  & 222 & 10.81 \\
TreesLindenSapling\_B  & 238 & 11.59 \\
TreesLindenSapling\_C  & 210 & 10.23 \\
\bottomrule
\end{tabular}
}
\label{tab:deciduous3_tree_mesh_composition}
\end{table}

This sampling approach can be applied to a specific region of interest and is compatible with our synthetic data generation pipeline. 
In fact, some laser scan data sets, such as the FOR-Species20K dataset \citep{puliti2024benchmarking}, have recently been made available to derive such individual tree meshes. 
Alternatively, high-fidelity reconstruction techniques such as 3D Gaussian Splatting \citep{kerbl20233d} can be experimented with to acquire high-fidelity tree assets with RGB cameras. 
In either approach, sampling individual trees for procedural foliage generation would be more efficient and reusable than traditional field work. 
The generated tree assets will be compatible with the data generation pipeline as presented in this paper. In addition, the generation of labels for machine learning would be cost-efficient.

\subsection{Further work}\label{sec:advance_real_forest_vision}

We provide some recommendations for future research aimed at developing robust representations for forest vision tasks using our synthetic data pipeline and the full-scale synthetic dataset generated via LiDAR simulation (Sim\textsubscript{11,187}).

\textbf{Mixed training with synthetic data when the target domain has no labelled data available }~\citep{dong2024towards,hatano2024multimodal}.
In addition to the scenario of \cref{sec:mixed_training}, in the pretraining phase, the standard supervision loss can be applied to $\mathcal{S}$ to learn task-specific features, for example segmentation. 
We then propose to incorporate a domain adaptation loss, such as those used in prior work on multimodal video data \citep{dong2024towards} to align representations between $\mathcal{S}$ and $\mathcal{T}_{\text{unlabel}}$. In that way, the pretrained representation can be better aligned to a specific region of interest.

\textbf{Scaling to neural architectures of higher capacity}
is another important aspect to consider. Although our study uses the algorithm from \citet{xiang2024automated} with its 3D convolutional neural network as backbone, Transformer architectures have shown promising results in point cloud processing \citep{wu2024point} and have recently been successfully applied to tree segmentation tasks \citep{xiang2025forestformer3d}. 
Empirical scaling laws for deep learning \citep{zhai2022scaling} show that Transformer architectures often continue to improve with increasing quantities of data, without clear saturation. Such models could especially benefit from using our synthetic data learn even stronger representations. 

\textbf{Expanding synthetic data generation} 
using our pipeline. 
For forest scenes from Unreal Engine, physics-based LiDAR simulation can be applied to not only UAV laser scans but also terrestrial, mobile, and airborne laser scans. 
Scene diversity can be enhanced by creating region-specific scenes, as noted in \cref{sec:efficient_field_sampling}. 
Unlimited variations of forest scenes can be generated by altering random seeds in procedural foliage generation. 
Currently, our experiments run with only two semantic labels, compared to five in the real dataset, which leads to some performance loss. This is a limitation of our generated data, and so is the best that is currently possible with few shot learning. 
Future work could explore generating finer semantic labels from Unreal Engine's tree assets to assess potential performance improvements.

To partially mitigate this limitation, we have extended the dataset to include leaf-wood semantic labels (\cref{fig:methods_synthetic_data_leafwood} in \cref{sec:extended_dataset_leafwood_appx}). This represents the most fine-grained annotation achievable with most existing tree assets designed for gaming applications. In particular, leaf–wood separation enables a key ecological application \citep{disney2018weighing}: fitting Quantitative Structure Models (QSMs) from wood points of individual trees to study branching patterns and forest growth dynamics. 
We have open-sourced our data pipeline to allow the community to enhance and contribute new forest scenes and laser scans, aiding in the advancement of forest vision systems.

\textbf{Evaluation benchmark:}
In addition to model training, our synthetic data could serve as a benchmark to evaluate the performance of different tree segmentation algorithms. This is possible because of its more extensive and balanced coverage across forest types.

%% file: 5_conclusion.tex
\section{Conclusion}
\label{sec:conclusions}
We have introduced a novel synthetic data generation pipeline for 3D forest vision to address a key challenge in this domain--- the scarcity of labelled data for supervised machine learning. 
Procedural foliage generation originated in the game industry but is rooted in ecological theories and that may partly explain its effectiveness. 
Using extensive tree assets from game engines, we can generate large-scale, diverse forest scenes. Combined with advanced LiDAR simulation, this enables a fully automated data generation pipeline to create machine learning-ready forest datasets. This facility supports a compelling methodology for more efficient field campaigns.

We used the pipeline to create an extensive 3D forest dataset that covers 12 scenes and 75 hectares, exceeding the size of any real dataset in this field that we have encountered so far. Comprehensive experiments confirmed the efficacy of the synthetic dataset for learning representations for instance segmentation. We have also highlighted three key elements that underpin the success of synthetic data: physics-based data generation, scene diversity, and a large scale dataset.
We hope and expect that researchers will be able to take advantage of our data pipeline and  generated data set to further advance forest vision using synthetic data. 

\paragraph{Acknowledgements:}
This work was supported by the UKRI Centre for Doctoral Training in Application of Artificial Intelligence to the study of Environmental Risks (reference EP/S022961/1) and Cambridge Centre for Carbon Credits.

%% file: 6_appendix.tex
\section{Appendix}\label{sec:appx}
\subsection{Key parameters used in the procedural foliage generation algorithm}\label{sec:procedural_parameters_appx}
Specifically, in each simulation: \texttt{Initial Seed Density} is the number of seeds placed along a 10-meter line. For example, a value of 3.0 means that $3^2 = 9$ seeds are placed per 10\,m $\times$ 10\,m area, approximately 900 seeds per hectare. 
\texttt{Number of Steps} is the number of simulation cycles used to spawn and thin the foliage.
\texttt{Average Spread Distance} is the mean distance by which new seeds are dispersed from a parent tree in each cycle, 
and \texttt{Spread Variance} adds randomness to the spread distance for more natural variability.

In addition, another group of parameters controls the structure of the forest after spawning seeds: \texttt{Collision Radius} is the minimum distance required between the foliage instances to prevent overlap. This can be interpreted as the space needed for a tree to survive, depending on its size.
\texttt{Shade Radius} is the area around each tree where its canopy suppresses the growth of new seedlings due to shading, 
and \texttt{Max Age} is the maximum number of simulation steps a foliage instance can endure before being removed. 
Furthermore, the foliage generation can be done for different groups of trees, e.g. overstory and understory, for a fine-grained modelling of the forest structures. 

\subsection{Supplementary metrics for \cref{sec:few_shot_learning_boost}}\label{sec:few_shot_learning_boost_appx}
Pretraining on our synthetic data allows the algorithm to achieve competitive performance after fine-tuning on only a few real forest plots, consistent across both coniferous (\cref{tab:eval_metrics_fine_tune_coniferous_appx}) and deciduous forest types (\cref{tab:eval_metrics_fine_tune_deciduous_appx}). With less than 0.1 ha of real data (the Real-C-S\textsubscript{1,1}), which represents only 2.2\% of the 2.79 ha FOR-Instance dataset, the model pretrained on our synthetic data is still able to achieve competitive performance.

\begin{table}[htbp!]
\centering
\caption{\textbf{The algorithm pretrained on our synthetic data can achieve competitive performance, with fine-tuning only on a few real forest plots.} \textit{This table presents the results for coniferous regions. The improvement of F1 score is significant compared to training on these few-shot dataset from scratch, indicating that our synthetic data contains useful prior knowledge for the algorithm to learn in order to generalise. All values are in percentages (\%).}}
\label{tab:eval_metrics_fine_tune_coniferous_appx}
\resizebox{\textwidth}{!}{
\begin{tabular}{lllcccc}
\hline
\textbf{Test} & \textbf{Pretrain} & \textbf{Train/Tune} & \textbf{Mean IoU} & \textbf{Precision} & \textbf{Recall} & \textbf{F1-score} \\ \hline

\multirow{4}{*}{Coniferous-N}  & - & Real\textsubscript{6,42} & 68.7 & 91.4 & 73.6 & 79.8 \\
                        & - & Real-C\textsubscript{3,3} & 39.2 & 80.2 & 40.2 & 50.5 \\
                        & Sim\textsubscript{11,187} & Real-C\textsubscript{3,3} & 67.1 & 94.8 & 71.1 & 79.4 \\
                        & Sim\textsubscript{11,187} & Real-C-N\textsubscript{1,1} & 66.0 & 91.3 & 69.4 & 77.1 \\\hline

\multirow{4}{*}{Coniferous-C}  & - & Real\textsubscript{6,42} & 88.3 & 95.1 & 93.0 & 93.4 \\
                        & - & Real-C\textsubscript{3,3} & 59.7 & 98.5 & 68.7 & 79.5 \\
                        & Sim\textsubscript{11,187} & Real-C\textsubscript{3,3} & 79.5 & 99.5 & 81.6 & 88.1 \\
                        & Sim\textsubscript{11,187} & Real-C-C\textsubscript{1,1} & 78.0 & 99.3 & 79.6 & 87.1 \\\hline

\multirow{4}{*}{Coniferous-S}  & - & Real\textsubscript{6,42} & 68.6 & 87.1 & 75.2 & 78.9 \\
                        & - & Real-C\textsubscript{3,3} & 54.4 & 91.4 & 59.1 & 69.2 \\
                        & Sim\textsubscript{11,187} & Real-C\textsubscript{3,3} & 67.4 & 91.7 & 75.4 & 81.4 \\
                        & Sim\textsubscript{11,187} & Real-C-S\textsubscript{1,1} & 61.1 & 92.7 & 66.9 & 75.7 \\\hline

\end{tabular}
}
\end{table}

\begin{table}[htbp!]
\centering
\caption{\textbf{The algorithm pretrained on our synthetic data can achieve competitive performance in deciduous regions, with fine-tuning on only a few real forest plots.} \textit{The pattern of improvement is similar to coniferous plots. These results strongly suggest that our synthetic data contain prior knowledge that can make the algorithm generalise for different forest types. All values are in percentages (\%).}}
\label{tab:eval_metrics_fine_tune_deciduous_appx}
\resizebox{\textwidth}{!}{
\begin{tabular}{lllcccc}
\hline
\textbf{Test Set} & \textbf{Pretrain} & \textbf{Train/Tune} & \textbf{Mean IoU} & \textbf{Precision} & \textbf{Recall} & \textbf{F1-score} \\ \hline
\multirow{4}{*}{Deciduous-R}  & - & Real\textsubscript{6,42} & 56.0 & 86.6 & 60.9 & 69.9 \\
                        & - & Real-D\textsubscript{2,2} & 27.1 & 67.1 & 22.4 & 31.6 \\
                        & Sim\textsubscript{11,187} & Real-D\textsubscript{2,2} & 61.5 & 81.7 & 63.7 & 70.0 \\
                        & Sim\textsubscript{11,187} & Real-D-R\textsubscript{1,1} & 61.8 & 82.1 & 62.9 & 69.3 \\\hline

\multirow{4}{*}{Deciduous-T}  & - & Real\textsubscript{6,42} & 48.3  & 63.8 & 48.8 & 52.5 \\
                        & - & Real-D\textsubscript{2,2} & 29.9  & 52.1 & 30.7 & 32.7 \\
                        & Sim\textsubscript{11,187} & Real-D\textsubscript{2,2} & 54.6 & 73.8 & 58.0 & 59.9 \\
                        & Sim\textsubscript{11,187} & Real-D-T\textsubscript{1,1} & 49.0 & 76.4 & 49.4 & 54.9 \\\hline

\end{tabular}
}
\end{table}

\FloatBarrier
\subsection{Supplementary metrics for \cref{sec:ablation_key_factors}}\label{sec:ablation_key_factors_appx}
Training with tree mesh point clouds (Nodal\textsubscript{4,20}) performs poorly, whereas UAV-LiDAR simulation (Sim\textsubscript{4,20}) significantly enhances performance (\cref{tab:eval_metrics_pretrain_ablations_appx}): 
F1 scores improve by over 30\% for Coniferous-C, Coniferous-S, and Deciduous-R, and over 20\% for other scenes. 

In particular, the zero-shot performance of Sim\textsubscript{11,187} exceeds training from scratch on few-shot datasets Real-C\textsubscript{3,3} and Real-D\textsubscript{2,2} for all scenes except Coniferous-S, where it is slightly worse (\cref{tab:eval_metrics_fine_tune_coniferous_appx}, \cref{tab:eval_metrics_fine_tune_deciduous_appx}). 

\begin{table}[htbp!]
\centering
\caption{\textbf{Three factors plays significant roles in generating strong synthetic data from our pipeline}, as revealed in the zero-shot experiment. \textit{(1) Physics-based simulation, by comparing improvement of F1-score from the algorithm trained on LiDAR simulation (Sim\textsubscript{4,20}) against nodal points from tree meshes Nodal\textsubscript{4,20}. (2) More diverse scenes, by comparing the improvement from training on Sim\textsubscript{4,20} against Sim\textsubscript{11,55}. Both datasets have five plots for each scene. (3) More data, by comparing the learning outcome of Sim\textsubscript{11,187} and Sim\textsubscript{11,55}. Both have the same number of scenes. All values are in percentages (\%).}}
\label{tab:eval_metrics_pretrain_ablations_appx}
\resizebox{\textwidth}{!}{
\begin{tabular}{lllcccc}
\hline
\textbf{Test} & \textbf{Train} & \textbf{Ablation Factor} & \textbf{Mean IoU} & \textbf{Precision} & \textbf{Recall} & \textbf{F1-score} \\ \hline

\multirow{4}{*}{Coniferous-N} & Nodal\textsubscript{4,20} & nodal points & 12.5 & 4.4 & 11.4 & 6.0 \\
                        & Sim\textsubscript{4,20} & LiDAR simulation & 29.0 & 37.0 & 28.6 & 28.4 \\
                        & Sim\textsubscript{11,55} & + diverse scenes & 44.9 & 69.0 & 45.1 & 50.8 \\
                        & Sim\textsubscript{11,187} & + more data & 50.3 & 71.5 & 56.8 & 59.4 \\\hline

\multirow{4}{*}{Coniferous-C}  & Nodal\textsubscript{4,20} & nodal points & 20.7 & 3.5 & 15.5 & 5.4 \\
                        & Sim\textsubscript{4,20} & LiDAR simulation & 38.6 & 63.5 & 34.3 & 42.6 \\
                        & Sim\textsubscript{11,55} & + diverse scenes & 72.4 & 95.7 & 74.0 & 81.5 \\
                        & Sim\textsubscript{11,187} & + more data & 81.2 & 96.6 & 81.6 & 87.0 \\\hline

\multirow{4}{*}{Coniferous-S}  & Nodal\textsubscript{4,20} & nodal points & 28.5 & 11.7 & 21.0 & 14.0 \\
                        & Sim\textsubscript{4,20} & LiDAR simulation & 44.0 & 70.7 & 46.2 & 52.4 \\
                        & Sim\textsubscript{11,55} & + diverse scenes & 46.2 & 63.7 & 49.9 & 53.0 \\
                        & Sim\textsubscript{11,187} & + more data & 51.9 & 59.9 & 55.8 & 55.9 \\\hline
                        
\multirow{4}{*}{Deciduous-R}  & Nodal\textsubscript{4,20} & nodal points & 16.9 & 2.9 & 5.8 & 2.7 \\
                        & Sim\textsubscript{4,20} & LiDAR simulation & 34.7 & 44.2 & 32.5 & 34.8 \\
                        & Sim\textsubscript{11,55} & + diverse scenes & 50.4 & 66.1 & 51.1 & 54.5 \\
                        & Sim\textsubscript{11,187} & + more data & 54.8 & 74.5 & 55.6 & 61.7 \\\hline

\multirow{4}{*}{Deciduous-T}  & Nodal\textsubscript{4,20} & nodal points & 20.1 & 3.7 & 16.8 & 5.2 \\
                        & Sim\textsubscript{4,20} & LiDAR simulation & 39.2 & 26.5 & 42.6 & 29.0 \\
                        & Sim\textsubscript{11,55} & + diverse scenes & 45.7 & 40.9 & 49.3 & 40.6 \\
                        & Sim\textsubscript{11,187} & + more data & 49.4 & 49.8 & 52.8 & 45.7 \\\hline

\end{tabular}
}
\end{table}

\subsection{Extended dataset with leaf--wood labels}\label{sec:extended_dataset_leafwood_appx}
\begin{figure}[htbp]
    \centering
    \includegraphics[width=\textwidth]{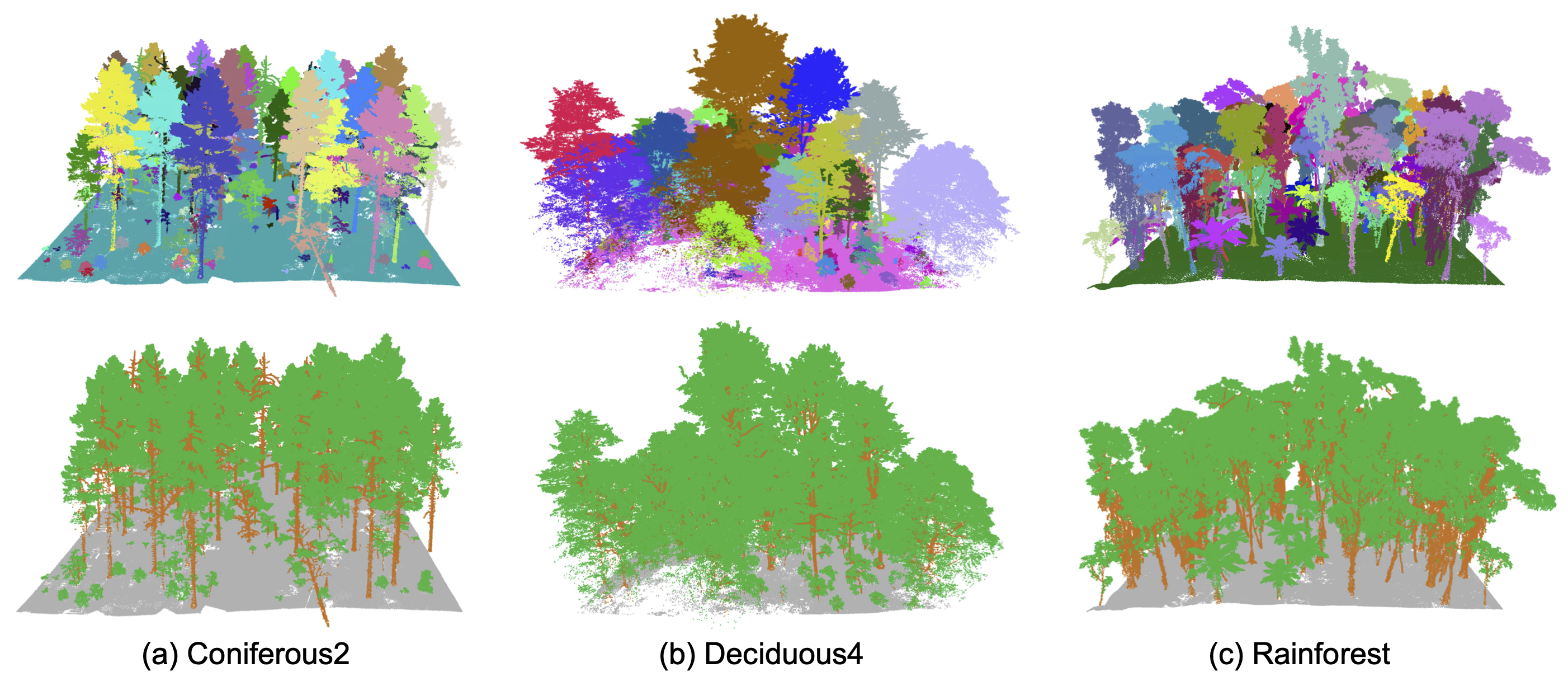}
    \caption{\textbf{Extended dataset with leaf--wood labels enabling quantitative analysis of forest structure.} \textit{Instance labels (top) and semantic labels (bottom) for ground, leaf, and wood, which are typically labor-intensive to generate through manual annotation.}}
    \label{fig:methods_synthetic_data_leafwood}
\end{figure}

\Cref{fig:methods_synthetic_data_leafwood} shows the extended dataset with leaf and wood labels, the most fine-grained annotations achievable with current tree assets. These labels enable the leaf--wood separation task, which future researchers can use to train forest vision systems for quantitative studies of forest structural dynamics. 

%% file: neurips_2024.bbl
\begin{thebibliography}{64}
\providecommand{\natexlab}[1]{#1}
\providecommand{\url}[1]{\texttt{#1}}
\expandafter\ifx\csname urlstyle\endcsname\relax
  \providecommand{\doi}[1]{doi: #1}\else
  \providecommand{\doi}{doi: \begingroup \urlstyle{rm}\Url}\fi

\bibitem[Swamy et~al.(2018)Swamy, Drazen, Johnson, and Bukoski]{swamy2018future}
Latha Swamy, Erika Drazen, Whitney~R Johnson, and Jacob~J Bukoski.
\newblock The future of tropical forests under the united nations sustainable development goals.
\newblock \emph{Journal of sustainable forestry}, 37\penalty0 (2):\penalty0 221--256, 2018.

\bibitem[Lines et~al.(2022)Lines, Fischer, Owen, and Jucker]{lines2022shape}
Emily~Rebecca Lines, Fabian~J{\"o}rg Fischer, Harry Jon~Foord Owen, and Tommaso Jucker.
\newblock The shape of trees: Reimagining forest ecology in three dimensions with remote sensing.
\newblock \emph{Journal of Ecology}, 110\penalty0 (8):\penalty0 1730--1745, 2022.

\bibitem[Liang et~al.(2022)Liang, Kukko, Balenovi{\'c}, Saarinen, Junttila, Kankare, Holopainen, Mokro{\v{s}}, Surov{\`y}, Kaartinen, et~al.]{liang2022close}
Xinlian Liang, Antero Kukko, Ivan Balenovi{\'c}, Ninni Saarinen, Samuli Junttila, Ville Kankare, Markus Holopainen, Martin Mokro{\v{s}}, Peter Surov{\`y}, Harri Kaartinen, et~al.
\newblock Close-range remote sensing of forests: The state of the art, challenges, and opportunities for systems and data acquisitions.
\newblock \emph{IEEE Geoscience and Remote Sensing Magazine}, 10\penalty0 (3):\penalty0 32--71, 2022.

\bibitem[Dubayah et~al.(2020)Dubayah, Blair, Goetz, Fatoyinbo, Hansen, Healey, Hofton, Hurtt, Kellner, Luthcke, et~al.]{dubayah2020global}
Ralph Dubayah, James~Bryan Blair, Scott Goetz, Lola Fatoyinbo, Matthew Hansen, Sean Healey, Michelle Hofton, George Hurtt, James Kellner, Scott Luthcke, et~al.
\newblock The global ecosystem dynamics investigation: High-resolution laser ranging of the earth’s forests and topography.
\newblock \emph{Science of remote sensing}, 1:\penalty0 100002, 2020.

\bibitem[Lang et~al.(2022)Lang, Kalischek, Armston, Schindler, Dubayah, and Wegner]{lang2022global}
Nico Lang, Nikolai Kalischek, John Armston, Konrad Schindler, Ralph Dubayah, and Jan~Dirk Wegner.
\newblock Global canopy height regression and uncertainty estimation from gedi lidar waveforms with deep ensembles.
\newblock \emph{Remote Sensing of Environment}, 268:\penalty0 112760, 2022.

\bibitem[Coomes et~al.(2017)Coomes, Dalponte, Jucker, Asner, Banin, Burslem, Lewis, Nilus, Phillips, Phua, et~al.]{coomes2017area}
David~A Coomes, Michele Dalponte, Tommaso Jucker, Gregory~P Asner, Lindsay~F Banin, David~FRP Burslem, Simon~L Lewis, Reuben Nilus, Oliver~L Phillips, Mui-How Phua, et~al.
\newblock Area-based vs tree-centric approaches to mapping forest carbon in southeast asian forests from airborne laser scanning data.
\newblock \emph{Remote Sensing of Environment}, 194:\penalty0 77--88, 2017.

\bibitem[Puliti et~al.(2024)Puliti, Lines, M{\"u}llerov{\'a}, Frey, Schindler, Straker, Allen, Winiwarter, Rehush, Hristova, et~al.]{puliti2024benchmarking}
Stefano Puliti, Emily~R Lines, Jana M{\"u}llerov{\'a}, Julian Frey, Zoe Schindler, Adrian Straker, Matthew~J Allen, Lukas Winiwarter, Nataliia Rehush, Hristina Hristova, et~al.
\newblock Benchmarking tree species classification from proximally-sensed laser scanning data: introducing the for-species20k dataset.
\newblock \emph{arXiv preprint arXiv:2408.06507}, 2024.

\bibitem[Chan et~al.(2021)Chan, Barnes, Swinfield, and Coomes]{chan2021monitoring}
Aland~HY Chan, Chloe Barnes, Tom Swinfield, and David~A Coomes.
\newblock Monitoring ash dieback (hymenoscyphus fraxineus) in british forests using hyperspectral remote sensing.
\newblock \emph{Remote Sensing in Ecology and Conservation}, 7\penalty0 (2):\penalty0 306--320, 2021.

\bibitem[Liang et~al.(2019)Liang, Wang, Py{\"o}r{\"a}l{\"a}, Lehtom{\"a}ki, Yu, Kaartinen, Kukko, Honkavaara, Issaoui, Nevalainen, et~al.]{liang2019forest}
Xinlian Liang, Yunsheng Wang, Jiri Py{\"o}r{\"a}l{\"a}, Matti Lehtom{\"a}ki, Xiaowei Yu, Harri Kaartinen, Antero Kukko, Eija Honkavaara, Aimad~EI Issaoui, Olli Nevalainen, et~al.
\newblock Forest in situ observations using unmanned aerial vehicle as an alternative of terrestrial measurements.
\newblock \emph{Forest ecosystems}, 6\penalty0 (1):\penalty0 1--16, 2019.

\bibitem[Chave et~al.(2009)Chave, Coomes, Jansen, Lewis, Swenson, and Zanne]{chave2009towards}
Jerome Chave, David Coomes, Steven Jansen, Simon~L Lewis, Nathan~G Swenson, and Amy~E Zanne.
\newblock Towards a worldwide wood economics spectrum.
\newblock \emph{Ecology letters}, 12\penalty0 (4):\penalty0 351--366, 2009.

\bibitem[Purves et~al.(2007)Purves, Lichstein, and Pacala]{purves2007crown}
Drew~W Purves, Jeremy~W Lichstein, and Stephen~W Pacala.
\newblock Crown plasticity and competition for canopy space: a new spatially implicit model parameterized for 250 north american tree species.
\newblock \emph{PloS one}, 2\penalty0 (9):\penalty0 e870, 2007.

\bibitem[Gauci et~al.(2024)Gauci, Pangala, Shenkin, Barba, Bastviken, Figueiredo, Gomez, Enrich-Prast, Sayer, Stauffer, et~al.]{gauci2024global}
Vincent Gauci, Sunitha~Rao Pangala, Alexander Shenkin, Josep Barba, David Bastviken, Viviane Figueiredo, Carla Gomez, Alex Enrich-Prast, Emma Sayer, Tain{\'a} Stauffer, et~al.
\newblock Global atmospheric methane uptake by upland tree woody surfaces.
\newblock \emph{Nature}, 631\penalty0 (8022):\penalty0 796--800, 2024.

\bibitem[Straker et~al.(2023)Straker, Puliti, Breidenbach, Kleinn, Pearse, Astrup, and Magdon]{straker2023instance}
Adrian Straker, Stefano Puliti, Johannes Breidenbach, Christoph Kleinn, Grant Pearse, Rasmus Astrup, and Paul Magdon.
\newblock Instance segmentation of individual tree crowns with yolov5: A comparison of approaches using the forinstance benchmark lidar dataset.
\newblock \emph{ISPRS Open Journal of Photogrammetry and Remote Sensing}, 9:\penalty0 100045, 2023.

\bibitem[Ball et~al.(2022)Ball, Hickman, Jackson, Koay, Hirst, Jay, Aubry-Kientz, Vincent, and Coomes]{ball2022accurate}
James~GC Ball, Sebastian~HM Hickman, Tobias~D Jackson, Xian~Jing Koay, James Hirst, William Jay, M{\'e}laine Aubry-Kientz, Gr{\'e}goire Vincent, and David~A Coomes.
\newblock Accurate tropical forest individual tree crown delineation from aerial rgb imagery using mask r-cnn.
\newblock \emph{bioRxiv}, pages 2022--07, 2022.

\bibitem[Wielgosz et~al.(2023)Wielgosz, Puliti, Wilkes, and Astrup]{wielgosz2023point2tree}
Maciej Wielgosz, Stefano Puliti, Phil Wilkes, and Rasmus Astrup.
\newblock Point2tree (p2t)—framework for parameter tuning of semantic and instance segmentation used with mobile laser scanning data in coniferous forest.
\newblock \emph{Remote Sensing}, 15\penalty0 (15):\penalty0 3737, 2023.

\bibitem[Xiang et~al.(2024)Xiang, Wielgosz, Kontogianni, Peters, Puliti, Astrup, and Schindler]{xiang2024automated}
Binbin Xiang, Maciej Wielgosz, Theodora Kontogianni, Torben Peters, Stefano Puliti, Rasmus Astrup, and Konrad Schindler.
\newblock Automated forest inventory: analysis of high-density airborne lidar point clouds with 3d deep learning.
\newblock \emph{Remote Sensing of Environment}, 305:\penalty0 114078, 2024.

\bibitem[Wielgosz et~al.(2024)Wielgosz, Puliti, Xiang, Schindler, and Astrup]{wielgosz2024segmentanytree}
Maciej Wielgosz, Stefano Puliti, Binbin Xiang, Konrad Schindler, and Rasmus Astrup.
\newblock Segmentanytree: A sensor and platform agnostic deep learning model for tree segmentation using laser scanning data.
\newblock \emph{arXiv preprint arXiv:2401.15739}, 2024.

\bibitem[Sutton(2019)]{sutton2019bitter}
Richard Sutton.
\newblock The bitter lesson.
\newblock \emph{Incomplete Ideas (blog)}, 13\penalty0 (1):\penalty0 38, 2019.

\bibitem[Weiser et~al.(2022)Weiser, Sch{\"a}fer, Winiwarter, Kra{\v{s}}ovec, Fassnacht, and H{\"o}fle]{weiser2022individual}
Hannah Weiser, Jannika Sch{\"a}fer, Lukas Winiwarter, Nina Kra{\v{s}}ovec, Fabian~E Fassnacht, and Bernhard H{\"o}fle.
\newblock Individual tree point clouds and tree measurements from multi-platform laser scanning in german forests.
\newblock \emph{Earth System Science Data}, 14\penalty0 (7):\penalty0 2989--3012, 2022.

\bibitem[Calders et~al.(2022)Calders, Verbeeck, Burt, Origo, Nightingale, Malhi, Wilkes, Raumonen, Bunce, and Disney]{calders2022laser}
Kim Calders, Hans Verbeeck, Andrew Burt, Niall Origo, Joanne Nightingale, Yadvinder Malhi, Phil Wilkes, Pasi Raumonen, Robert~GH Bunce, and Mathias Disney.
\newblock Laser scanning reveals potential underestimation of biomass carbon in temperate forest.
\newblock \emph{Ecological Solutions and Evidence}, 3\penalty0 (4):\penalty0 e12197, 2022.

\bibitem[Puliti et~al.(2023)Puliti, Pearse, Surov{\`y}, Wallace, Hollaus, Wielgosz, and Astrup]{puliti2023instance}
Stefano Puliti, Grant Pearse, Peter Surov{\`y}, Luke Wallace, Markus Hollaus, Maciej Wielgosz, and Rasmus Astrup.
\newblock For-instance: a uav laser scanning benchmark dataset for semantic and instance segmentation of individual trees.
\newblock \emph{arXiv preprint arXiv:2309.01279}, 2023.

\bibitem[Uddin and Lu(2024)]{uddin2024dataset}
Shahadat Uddin and Haohui Lu.
\newblock Dataset meta-level and statistical features affect machine learning performance.
\newblock \emph{Scientific Reports}, 14\penalty0 (1):\penalty0 1670, 2024.

\bibitem[Shotton et~al.(2011)Shotton, Fitzgibbon, Cook, Sharp, Finocchio, Moore, Kipman, and Blake]{shotton2011real}
Jamie Shotton, Andrew Fitzgibbon, Mat Cook, Toby Sharp, Mark Finocchio, Richard Moore, Alex Kipman, and Andrew Blake.
\newblock Real-time human pose recognition in parts from single depth images.
\newblock In \emph{CVPR 2011}, pages 1297--1304. Ieee, 2011.

\bibitem[Sklyarova et~al.(2023)Sklyarova, Zakharov, Hilliges, Black, and Thies]{sklyarova2023haar}
Vanessa Sklyarova, Egor Zakharov, Otmar Hilliges, Michael~J Black, and Justus Thies.
\newblock Haar: Text-conditioned generative model of 3d strand-based human hairstyles.
\newblock \emph{arXiv preprint arXiv:2312.11666}, 2023.

\bibitem[Gaidon et~al.(2016)Gaidon, Wang, Cabon, and Vig]{gaidon2016virtual}
Adrien Gaidon, Qiao Wang, Yohann Cabon, and Eleonora Vig.
\newblock Virtual worlds as proxy for multi-object tracking analysis.
\newblock In \emph{Proceedings of the IEEE conference on computer vision and pattern recognition}, pages 4340--4349, 2016.

\bibitem[Roberts et~al.(2021)Roberts, Ramapuram, Ranjan, Kumar, Bautista, Paczan, Webb, and Susskind]{roberts2021hypersim}
Mike Roberts, Jason Ramapuram, Anurag Ranjan, Atulit Kumar, Miguel~Angel Bautista, Nathan Paczan, Russ Webb, and Joshua~M Susskind.
\newblock Hypersim: A photorealistic synthetic dataset for holistic indoor scene understanding.
\newblock In \emph{Proceedings of the IEEE/CVF international conference on computer vision}, pages 10912--10922, 2021.

\bibitem[Song et~al.(2023)Song, He, Li, Ma, Ming, Mao, Pei, Peng, Hu, Yao, et~al.]{song2023synthetic}
Zhihang Song, Zimin He, Xingyu Li, Qiming Ma, Ruibo Ming, Zhiqi Mao, Huaxin Pei, Lihui Peng, Jianming Hu, Danya Yao, et~al.
\newblock Synthetic datasets for autonomous driving: A survey.
\newblock \emph{IEEE Transactions on Intelligent Vehicles}, 9\penalty0 (1):\penalty0 1847--1864, 2023.

\bibitem[Holmberg(2025)]{holmberg2025unreal}
Nora Holmberg.
\newblock Unreal engine environment generation (pcg): a comparative overview of existing tools.
\newblock 2025.

\bibitem[Grondin et~al.(2022)Grondin, Fortin, Pomerleau, and Giguère]{Grondin_2022}
Vincent Grondin, Jean-Michel Fortin, François Pomerleau, and Philippe Giguère.
\newblock Tree detection and diameter estimation based on deep learning.
\newblock \emph{Forestry: An International Journal of Forest Research}, 96\penalty0 (2):\penalty0 264–276, October 2022.
\newblock ISSN 1464-3626.
\newblock \doi{10.1093/forestry/cpac043}.
\newblock URL \url{http://dx.doi.org/10.1093/forestry/cpac043}.

\bibitem[Lu et~al.(2024)Lu, Huang, Sun, Zhang, Zhang, Fei, and Chen]{lu2024m2fnet}
Yawen Lu, Yunhan Huang, Su~Sun, Tansi Zhang, Xuewen Zhang, Songlin Fei, and Victor Chen.
\newblock M2fnet: Multi-modal forest monitoring network on large-scale virtual dataset.
\newblock \emph{arXiv preprint arXiv:2402.04534}, 2024.

\bibitem[Feng et~al.(2025)Feng, She, and Keshav]{feng2025spread}
Zhengpeng Feng, Yihang She, and Srinivasan Keshav.
\newblock Spread: A large-scale, high-fidelity synthetic dataset for multiple forest vision tasks.
\newblock \emph{Ecological Informatics}, 87:\penalty0 103085, 2025.

\bibitem[Liu et~al.(2021)Liu, Tucker, Jampani, Makadia, Snavely, and Kanazawa]{liu2021infinite}
Andrew Liu, Richard Tucker, Varun Jampani, Ameesh Makadia, Noah Snavely, and Angjoo Kanazawa.
\newblock Infinite nature: Perpetual view generation of natural scenes from a single image.
\newblock In \emph{Proceedings of the IEEE/CVF International Conference on Computer Vision}, pages 14458--14467, 2021.

\bibitem[Winiwarter et~al.(2022)Winiwarter, Pena, Weiser, Anders, S{\'a}nchez, Searle, and H{\"o}fle]{winiwarter2022virtual}
Lukas Winiwarter, Alberto Manuel~Esmor{\'\i}s Pena, Hannah Weiser, Katharina Anders, Jorge~Mart{\'\i}nez S{\'a}nchez, Mark Searle, and Bernhard H{\"o}fle.
\newblock Virtual laser scanning with helios++: A novel take on ray tracing-based simulation of topographic full-waveform 3d laser scanning.
\newblock \emph{Remote Sensing of Environment}, 269:\penalty0 112772, 2022.

\bibitem[Wang(2020)]{wang2020unsupervised}
Di~Wang.
\newblock Unsupervised semantic and instance segmentation of forest point clouds.
\newblock \emph{ISPRS Journal of Photogrammetry and Remote Sensing}, 165:\penalty0 86--97, 2020.

\bibitem[Bryson et~al.(2024)Bryson, Ravendran, Mercier, Frickey, Jayathunga, Pearse, and Hartley]{bryson2024domain}
Mitch Bryson, Ahalya Ravendran, Celine Mercier, Tancred Frickey, Sadeepa Jayathunga, Grant Pearse, and Robin~JL Hartley.
\newblock Domain adaptation of deep neural networks for tree part segmentation using synthetic forest trees.
\newblock \emph{ISPRS Open Journal of Photogrammetry and Remote Sensing}, 14:\penalty0 100078, 2024.

\bibitem[Tobin et~al.(2017)Tobin, Fong, Ray, Schneider, Zaremba, and Abbeel]{tobin2017domain}
Josh Tobin, Rachel Fong, Alex Ray, Jonas Schneider, Wojciech Zaremba, and Pieter Abbeel.
\newblock Domain randomization for transferring deep neural networks from simulation to the real world.
\newblock In \emph{2017 IEEE/RSJ international conference on intelligent robots and systems (IROS)}, pages 23--30. IEEE, 2017.

\bibitem[Prakash et~al.(2019)Prakash, Boochoon, Brophy, Acuna, Cameracci, State, Shapira, and Birchfield]{prakash2019structured}
Aayush Prakash, Shaad Boochoon, Mark Brophy, David Acuna, Eric Cameracci, Gavriel State, Omer Shapira, and Stan Birchfield.
\newblock Structured domain randomization: Bridging the reality gap by context-aware synthetic data.
\newblock In \emph{2019 International Conference on Robotics and Automation (ICRA)}, pages 7249--7255. IEEE, 2019.

\bibitem[Finn et~al.(2017)Finn, Abbeel, and Levine]{finn2017model}
Chelsea Finn, Pieter Abbeel, and Sergey Levine.
\newblock Model-agnostic meta-learning for fast adaptation of deep networks.
\newblock In \emph{International conference on machine learning}, pages 1126--1135. PMLR, 2017.

\bibitem[Liu et~al.(2025)Liu, Wang, Gong, Wang, Zhu, and Wang]{liu2025advancing}
Jing Liu, Duanchu Wang, Haoran Gong, Chongyu Wang, Jihua Zhu, and Di~Wang.
\newblock Advancing the understanding of fine-grained 3d forest structures using digital cousins and simulation-to-reality: Methods and datasets.
\newblock \emph{arXiv preprint arXiv:2501.03637}, 2025.

\bibitem[Lang et~al.(2023)Lang, Jetz, Schindler, and Wegner]{lang2023high}
Nico Lang, Walter Jetz, Konrad Schindler, and Jan~Dirk Wegner.
\newblock A high-resolution canopy height model of the earth.
\newblock \emph{Nature Ecology \& Evolution}, pages 1--12, 2023.

\bibitem[Okura(2022)]{okura20223d}
Fumio Okura.
\newblock 3d modeling and reconstruction of plants and trees: A cross-cutting review across computer graphics, vision, and plant phenotyping.
\newblock \emph{Breeding Science}, 72\penalty0 (1):\penalty0 31--47, 2022.

\bibitem[Nesbit and Hugenholtz(2019)]{nesbit2019enhancing}
Paul~Ryan Nesbit and Christopher~H Hugenholtz.
\newblock Enhancing uav--sfm 3d model accuracy in high-relief landscapes by incorporating oblique images.
\newblock \emph{Remote Sensing}, 11\penalty0 (3):\penalty0 239, 2019.

\bibitem[Flynn(2024)]{flynn2024technological}
W~Flynn.
\newblock Technological advances in close range sensors and methodological complexities for measuring forest structure and disturbance.
\newblock 2024.

\bibitem[Xiang et~al.(2023)Xiang, Peters, Kontogianni, Vetterli, Puliti, Astrup, and Schindler]{xiang2023towards}
Binbin Xiang, Torben Peters, Theodora Kontogianni, Frawa Vetterli, Stefano Puliti, Rasmus Astrup, and Konrad Schindler.
\newblock Towards accurate instance segmentation in large-scale lidar point clouds.
\newblock \emph{arXiv preprint arXiv:2307.02877}, 2023.

\bibitem[Liang et~al.(2018)Liang, Hyypp{\"a}, Kaartinen, Lehtom{\"a}ki, Py{\"o}r{\"a}l{\"a}, Pfeifer, Holopainen, Brolly, Francesco, Hackenberg, et~al.]{liang2018international}
Xinlian Liang, Juha Hyypp{\"a}, Harri Kaartinen, Matti Lehtom{\"a}ki, Jiri Py{\"o}r{\"a}l{\"a}, Norbert Pfeifer, Markus Holopainen, G{\'a}bor Brolly, Pirotti Francesco, Jan Hackenberg, et~al.
\newblock International benchmarking of terrestrial laser scanning approaches for forest inventories.
\newblock \emph{ISPRS journal of photogrammetry and remote sensing}, 144:\penalty0 137--179, 2018.

\bibitem[Calders et~al.(2016)Calders, Burt, Origo, Disney, Nightingale, Raumonen, and Lewis]{calders2016large}
Kim Calders, Andrew Burt, Niall Origo, Mathias Disney, J~Nightingale, Pasi Raumonen, and Philip Lewis.
\newblock Large-area virtual forests from terrestrial laser scanning data.
\newblock In \emph{2016 IEEE international geoscience and remote sensing symposium (IGARSS)}, pages 1765--1767. IEEE, 2016.

\bibitem[Dosovitskiy et~al.(2017)Dosovitskiy, Ros, Codevilla, Lopez, and Koltun]{dosovitskiy2017carla}
Alexey Dosovitskiy, German Ros, Felipe Codevilla, Antonio Lopez, and Vladlen Koltun.
\newblock Carla: An open urban driving simulator.
\newblock In \emph{Conference on robot learning}, pages 1--16. PMLR, 2017.

\bibitem[Bondi et~al.(2018)Bondi, Dey, Kapoor, Piavis, Shah, Fang, Dilkina, Hannaford, Iyer, Joppa, et~al.]{bondi2018airsim}
Elizabeth Bondi, Debadeepta Dey, Ashish Kapoor, Jim Piavis, Shital Shah, Fei Fang, Bistra Dilkina, Robert Hannaford, Arvind Iyer, Lucas Joppa, et~al.
\newblock Airsim-w: A simulation environment for wildlife conservation with uavs.
\newblock In \emph{Proceedings of the 1st ACM SIGCAS Conference on Computing and Sustainable Societies}, pages 1--12, 2018.

\bibitem[Esmor{\'\i}s et~al.(2024)Esmor{\'\i}s, Weiser, Winiwarter, Cabaleiro, and H{\"o}fle]{esmoris2024deep}
Alberto~M Esmor{\'\i}s, Hannah Weiser, Lukas Winiwarter, Jose~C Cabaleiro, and Bernhard H{\"o}fle.
\newblock Deep learning with simulated laser scanning data for 3d point cloud classification.
\newblock \emph{ISPRS Journal of Photogrammetry and Remote Sensing}, 215:\penalty0 192--213, 2024.

\bibitem[Sch{\"a}fer et~al.(2023)Sch{\"a}fer, Weiser, Winiwarter, H{\"o}fle, Schmidtlein, and Fassnacht]{schafer2023generating}
Jannika Sch{\"a}fer, Hannah Weiser, Lukas Winiwarter, Bernhard H{\"o}fle, Sebastian Schmidtlein, and Fabian~Ewald Fassnacht.
\newblock Generating synthetic laser scanning data of forests by combining forest inventory information, a tree point cloud database and an open-source laser scanning simulator.
\newblock \emph{Forestry: An International Journal of Forest Research}, 96\penalty0 (5):\penalty0 653--671, 2023.

\bibitem[Liu et~al.(2022)Liu, Calders, Meunier, Gastellu-Etchegorry, Nightingale, Disney, Origo, Woodgate, and Verbeeck]{liu2022implications}
Chang Liu, Kim Calders, F{\'e}licien Meunier, JP~Gastellu-Etchegorry, J~Nightingale, M~Disney, N~Origo, W~Woodgate, and Hans Verbeeck.
\newblock Implications of 3d forest stand reconstruction methods for radiative transfer modeling: A case study in the temperate deciduous forest.
\newblock \emph{Journal of Geophysical Research: Atmospheres}, 127\penalty0 (14):\penalty0 e2021JD036175, 2022.

\bibitem[Hall{\'e} and Oldeman(1970)]{halle1970essay}
Francis Hall{\'e} and Roelof~AA Oldeman.
\newblock Essay on the architecture and dynamics of growth of tropical trees.
\newblock 1970.

\bibitem[Pacala et~al.(1996)Pacala, Canham, Saponara, Silander~Jr, Kobe, and Ribbens]{pacala1996forest}
Stephen~W Pacala, Charles~D Canham, John Saponara, John~A Silander~Jr, Richard~K Kobe, and Eric Ribbens.
\newblock Forest models defined by field measurements: estimation, error analysis and dynamics.
\newblock \emph{Ecological monographs}, 66\penalty0 (1):\penalty0 1--43, 1996.

\bibitem[Neumann et~al.(2022)Neumann, Borrmann, and N{\"u}chter]{neumann2022semantic}
Michael Neumann, Dorit Borrmann, and Andreas N{\"u}chter.
\newblock Semantic classification in uncolored 3d point clouds using multiscale features.
\newblock In \emph{International Conference on Intelligent Autonomous Systems}, pages 342--359. Springer, 2022.

\bibitem[Jiang et~al.(2020)Jiang, Zhao, Shi, Liu, Fu, and Jia]{jiang2020pointgroup}
Li~Jiang, Hengshuang Zhao, Shaoshuai Shi, Shu Liu, Chi-Wing Fu, and Jiaya Jia.
\newblock Pointgroup: Dual-set point grouping for 3d instance segmentation.
\newblock In \emph{Proceedings of the IEEE/CVF conference on computer vision and Pattern recognition}, pages 4867--4876, 2020.

\bibitem[Choy et~al.(2019)Choy, Gwak, and Savarese]{choy20194d}
Christopher Choy, JunYoung Gwak, and Silvio Savarese.
\newblock 4d spatio-temporal convnets: Minkowski convolutional neural networks.
\newblock In \emph{Proceedings of the IEEE Conference on Computer Vision and Pattern Recognition}, pages 3075--3084, 2019.

\bibitem[McInnes et~al.(2018)McInnes, Healy, and Melville]{mcinnes2018umap}
Leland McInnes, John Healy, and James Melville.
\newblock Umap: Uniform manifold approximation and projection for dimension reduction.
\newblock \emph{arXiv preprint arXiv:1802.03426}, 2018.

\bibitem[Kerbl et~al.(2023)Kerbl, Kopanas, Leimk{\"u}hler, and Drettakis]{kerbl20233d}
Bernhard Kerbl, Georgios Kopanas, Thomas Leimk{\"u}hler, and George Drettakis.
\newblock 3d gaussian splatting for real-time radiance field rendering.
\newblock \emph{ACM Transactions on Graphics}, 42\penalty0 (4):\penalty0 1--14, 2023.

\bibitem[Dong et~al.(2024)Dong, Chatzi, and Fink]{dong2024towards}
Hao Dong, Eleni Chatzi, and Olga Fink.
\newblock Towards multimodal open-set domain generalization and adaptation through self-supervision.
\newblock In \emph{European Conference on Computer Vision}, pages 270--287. Springer, 2024.

\bibitem[Hatano et~al.(2024)Hatano, Hachiuma, Fujii, and Saito]{hatano2024multimodal}
Masashi Hatano, Ryo Hachiuma, Ryo Fujii, and Hideo Saito.
\newblock Multimodal cross-domain few-shot learning for egocentric action recognition.
\newblock In \emph{European Conference on Computer Vision}, pages 182--199. Springer, 2024.

\bibitem[Wu et~al.(2024)Wu, Jiang, Wang, Liu, Liu, Qiao, Ouyang, He, and Zhao]{wu2024point}
Xiaoyang Wu, Li~Jiang, Peng-Shuai Wang, Zhijian Liu, Xihui Liu, Yu~Qiao, Wanli Ouyang, Tong He, and Hengshuang Zhao.
\newblock Point transformer v3: Simpler faster stronger.
\newblock In \emph{Proceedings of the IEEE/CVF conference on computer vision and pattern recognition}, pages 4840--4851, 2024.

\bibitem[Xiang et~al.(2025)Xiang, Wielgosz, Puliti, Kr{\'a}l, Kr{\v{u}}{\v{c}}ek, Missarov, and Astrup]{xiang2025forestformer3d}
Binbin Xiang, Maciej Wielgosz, Stefano Puliti, Kamil Kr{\'a}l, Martin Kr{\v{u}}{\v{c}}ek, Azim Missarov, and Rasmus Astrup.
\newblock Forestformer3d: A unified framework for end-to-end segmentation of forest lidar 3d point clouds.
\newblock \emph{arXiv preprint arXiv:2506.16991}, 2025.

\bibitem[Zhai et~al.(2022)Zhai, Kolesnikov, Houlsby, and Beyer]{zhai2022scaling}
Xiaohua Zhai, Alexander Kolesnikov, Neil Houlsby, and Lucas Beyer.
\newblock Scaling vision transformers.
\newblock In \emph{Proceedings of the IEEE/CVF conference on computer vision and pattern recognition}, pages 12104--12113, 2022.

\bibitem[Disney et~al.(2018)Disney, Boni~Vicari, Burt, Calders, Lewis, Raumonen, and Wilkes]{disney2018weighing}
Mathias~I Disney, Matheus Boni~Vicari, Andrew Burt, Kim Calders, Simon~L Lewis, Pasi Raumonen, and Phil Wilkes.
\newblock Weighing trees with lasers: advances, challenges and opportunities.
\newblock \emph{Interface Focus}, 8\penalty0 (2):\penalty0 20170048, 2018.

\end{thebibliography}
